\pdfoutput=1

\documentclass[11pt]{article}

\usepackage{EMNLP2022}

\usepackage{times}
\usepackage{latexsym}
\usepackage{booktabs}

\usepackage[T1]{fontenc}

\usepackage[utf8]{inputenc}

\usepackage{microtype}

\usepackage{inconsolata}

\usepackage{CJKutf8} 

\usepackage{xcolor}

\usepackage{multirow}

\usepackage{graphicx}

\usepackage{tablefootnote}

\usepackage{makecell}

\definecolor{atomictangerine}{rgb}{1.0, 0.6, 0.4}

\newcommand{\name}[0]{\textsc{Demetr}}

\title{\includegraphics[scale=0.4]{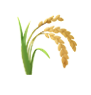}\name: Diagnosing Evaluation Metrics for Translation}

\author{Marzena Karpinska$^\diamondsuit$ \quad Nishant Raj$^\diamondsuit$ \quad Katherine Thai$^\diamondsuit$ \\  {\bf Yixiao Song$^\spadesuit$ \quad Ankita Gupta$^\diamondsuit$ \quad Mohit Iyyer$^\diamondsuit$}\\\\
$^\diamondsuit$Manning College of Information and Computer Sciences, UMass Amherst\\
$^\spadesuit$Department of Linguistics, UMass Amherst\\
\texttt{\{mkarpinska,kbthai,ankitagupta,miyyer\}@cs.umass.edu} \\ 
\texttt{\{nishantraj,yixiaosong\}@umass.edu}}

\begin{document}
\maketitle

\begin{abstract}
While machine translation evaluation metrics based on string overlap (e.g., \textsc{Bleu}) have their limitations, their computations are transparent: the \textsc{Bleu} score assigned to a particular candidate translation can be traced back to the presence or absence of certain words. The operations of newer \emph{learned} metrics (e.g., \textsc{Bleurt}, \textsc{Comet}), which leverage pretrained language models to achieve higher correlations with human quality judgments than \textsc{Bleu}, are opaque in comparison.  In this paper, we shed light on the behavior of these learned metrics by creating \name, a diagnostic dataset with \emph{31}K English examples (translated from \emph{10} source languages) for evaluating the sensitivity of MT evaluation metrics to \emph{35} different linguistic perturbations spanning semantic, syntactic, and morphological error categories.
All perturbations were carefully designed to form minimal pairs with the actual translation (i.e., differ in only one aspect). 
We find that learned metrics perform substantially better than string-based metrics on \name. 
Additionally, learned metrics differ in their sensitivity to various phenomena (e.g., \textsc{BERTScore} is sensitive to untranslated words but relatively insensitive to gender manipulation, while \textsc{Comet} is much more sensitive to word repetition than to aspectual changes).
We publicly release \name\ to spur more informed future development of machine translation evaluation metrics\footnote{\url{https://github.com/marzenakrp/demetr}}.
\end{abstract}
\section{Introduction}
\label{sec:introduction}

Automatically evaluating the output quality of machine translation (MT) systems remains a difficult challenge. The \textsc{Bleu} metric~\citep{papineni-etal-2002-bleu}, which is a function of $n$-gram overlap between system and reference outputs, is still used widely today despite its obvious limitations in measuring semantic similarity \citep{fomicheve-et-al-2019, marie-etal-2021-scientific, shipornotKomci, freitag-et-al-blue-surface}. Recently-developed \emph{learned} evaluation metrics such as \textsc{Bleurt} \citep{sellam-etal-2020-bleurt}, \textsc{Comet} \citep{rei-etal-2020-comet}, \textsc{MoverScore} \citep{zhao-etal-2019-moverscore}, or \textsc{BARTScore} \citep{bartscore} seek to address these limitations by either fine-tuning pretrained language models directly on human judgments of translation quality or by simply utilizing contextualized word embeddings. While learned metrics exhibit higher correlation with human judgments than \textsc{Bleu} \citep{wmt-2021-machine}, their relative lack of interpretability leaves it unclear as to \emph{why} they assign a particular score to a given translation. This is a major reason why some MT researchers are reluctant to employ learned metrics in order to evaluate their MT systems \citep{marie-etal-2021-scientific, repairing-cracked-evaluation, towards-explainable}. 

\begin{figure}[t]
    \centering
    \includegraphics[width=0.48\textwidth]{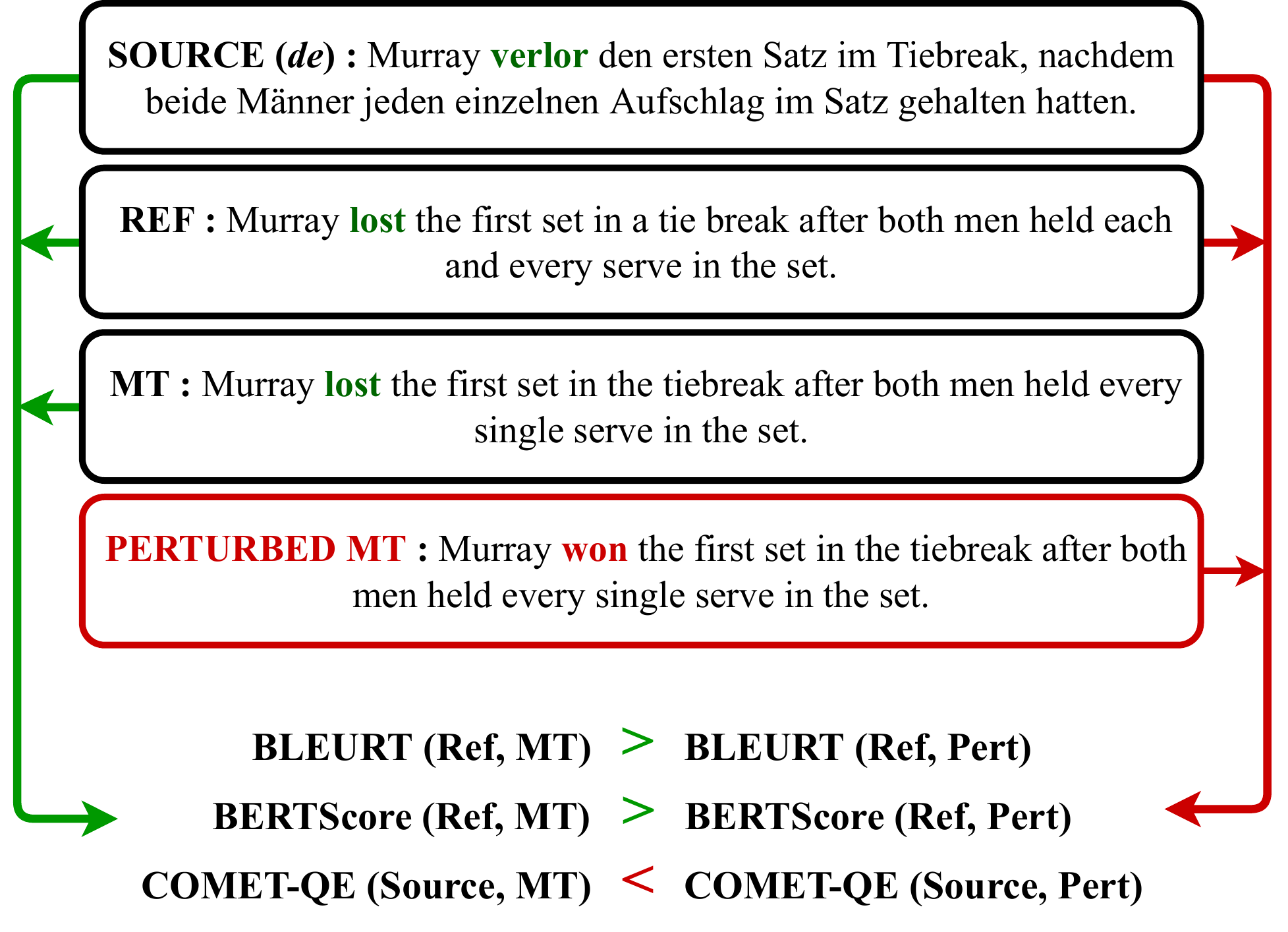}
    \caption{An example perturbation (antonym replacement) from our \name~dataset. We measure whether different MT evaluation metrics score the unperturbed translation higher than the perturbed translation; in this case, \textsc{Bleurt} and \textsc{BERTScore} accurately identify the perturbation, while \textsc{Comet-QE} fails to do so.}
    \label{fig:demeter-overview}
\end{figure}

In this paper, we build on previous metric explainability work \citep{specia-etal-2010-dataset, macketanz-etal-2018-tq, fomicheve-et-al-2019,  kaster-etal-2021-global, sai-etal-2021-perturbation, wmt-2021-machine, fomicheva-etal-2021-eval4nlp, towards-explainable} by introducing \name, a dataset for \textbf{D}iagnosing \textbf{E}valuation \textbf{METR}ics for machine translation, that measures the sensitivity of an MT metric to \emph{35} different types of linguistic perturbations spanning common syntactic (e.g., incorrect word order), semantic (e.g., undertranslation), and morphological (e.g., incorrect suffix) translation error categories. Each example in \name\ is a tuple containing \textbf{\{}\texttt{{\color{teal}\textbf{source}}, {\color{brown}\textbf{reference}},
{\color{violet}\textbf{machine translation}}, {\color{purple}\textbf{perturbed machine translation}}}\textbf{\}}, as shown in Figure~\ref{fig:demeter-overview}. The entire dataset contains of \emph{31}K total examples across \emph{10} different source languages (the target language is always English). The perturbations in \name\ are produced semi-automatically by manipulating translations produced by commercial MT systems such as Google Translate, and they are manually validated to ensure the only source of variation is associated with the desired perturbation. 

We measure the accuracy of a suite of \emph{14} evaluation metrics on \name\ (as shown in Figure~\ref{fig:demeter-overview}), discovering that learned metrics perform far better than string-based ones. We also analyze the relative \emph{sensitivity} of metrics to different grades of perturbation severity. We find that metrics struggle at times to differentiate between minor errors (e.g., punctuation removal or word repetition)  with semantics-warping errors such as incorrect gender or numeracy. We also observe that the reference-free\footnote{While prior work uses also terms such as ``reference-less'' and ``quality estimation,'' we employ the term ``reference-free" as it is more self-explanatory.} \textsc{Comet-QE} learned metric is more sensitive to word repetition and misspelled words than severe errors such as entirely unrelated translations or named entity replacement. 
We publicly release \name\ and associated code to facilitate more principled research into MT evaluation.

\section{Diagnosing MT evaluation metrics}
Most existing MT evaluation metrics compute a score for a candidate translation $t$ against a reference sentence $r$.\footnote{Some metrics, such as \textsc{Comet}, additionally condition the score on the source sentence.} These scores can be either a simple function of character or token overlap between $t$ and $r$ (e.g., \textsc{Bleu}), or they can be the result of a complex neural network model that embeds $t$ and $r$ (e.g., \textsc{Bleurt}). While the latter class of \emph{learned} metrics\footnote{We define \textit{learned} metrics as any metric which uses a machine learning model (including both pretrained and supervised methods).} provides more meaningful judgments of translation quality than the former, they are also relatively uninterpretable: the reason for a particular translation $t$ receiving a high or low score is difficult to discern. In this section, we first explain our perturbation-based methodology to better understand MT metrics before describing the collection of \name, a dataset of linguistic perturbations. 

\subsection{Using translation perturbations to diagnose MT metrics}\label{sec.using.translation.perturbations.to.diagnose.MT}
Inspired by prior work in \emph{minimal pair}-based linguistic evaluation of pretrained language models such as \textsc{BLiMP}~\citep{warstadt-etal-2020-blimp-benchmark}, we investigate how sensitive MT evaluation metrics are to various perturbations of the candidate translation $t$. Consider the following example, which is designed to evaluate the impact of word order in the candidate translation:

\begin{quote}
\small
    {\color{brown}\textbf{\texttt{reference translation}}} \textbf{$r$}: Pronunciation is relatively easy in Italian since most words are pronounced exactly how they are written.\vspace{0.1cm}\\
{\color{violet}\textbf{\texttt{machine translation}}} \textbf{$t$}: Pronunciation is relatively easy in Italian, as most words are pronounced exactly as they are spelled.\vspace{0.1cm}\\
{\color{purple}\textbf{\texttt{perturbed machine translation}}} \textbf{$t'$}: Spelled pronunciation as Italian, relatively are most is as they pronounced exactly in words easy.
\end{quote}

If a particular evaluation metric \texttt{SCORE} is sensitive to this shuffling perturbation, \texttt{SCORE}$(r, t')$, the score of the perturbed translation, should be lower than \texttt{SCORE}$(r, t)$.\footnote{For reference-free metrics like \textsc{Comet-QE}, we include the source sentence $s$ as an input to the scoring function instead of the reference.} Note that while other minor translation errors may be present in $t$, the perturbed translation $t'$ differs only in a specific, controlled perturbation (in this case, shuffling). 

\subsection{Creating the \name\ dataset}
To explore the above methodology at scale, we create \name, a dataset that evaluates MT metrics on \emph{35} different linguistic phenomena with 1K perturbations per phenomenon.\footnote{As some perturbations require presence of specific items (e.g., to omit a named entity, one has to be present) not all perturbations include exactly 1k sentences.}
Each example in \name\ consists of (1) a sentence in one of \emph{10} source languages, (2) an English translation written by a human translator, (3) a machine translation produced by Google Translate, \footnote{We edit the machine translation to assure a satisfactory quality. In cases where the Google Translate output is exceptionally poor, we either replace the sentence or replace the translation with one produced by \href{https://www.deepl.com/translator}{DeepL~\citep{frahling_n.d.}} or \textsc{GPT-3}~\citep{brown2020language}.} and (4) a perturbed version of the Google Translate output which introduces exactly one mistake (semantic, syntactic, or typographical). 

\begin{table}[t!]
\small
\begin{center}
\resizebox{\columnwidth}{!}{\begin{tabular}{cclc} 
 \toprule
\textbf{ID} & \textbf{Category} & \textbf{Description} & \textbf{Error severity} \\
 \midrule
  1 & \multirow{11}{*}{\rotatebox{90}{\textbf{accuracy}}} & word repetition (twice) & \textcolor{teal}{\textbf{minor}}\\
  2 & & word repetition (four times) & \textcolor{teal}{\textbf{minor}}\\
  3 & & too general word (undertranslation) & \textcolor{violet}{\textbf{major}}\\
  4 & & untranslated word (codemix) & \textcolor{violet}{\textbf{major}}\\
  5 & & omitted perpositional phrase & \textcolor{violet}{\textbf{major}}\\
  6 & & incorrect word added & \textcolor{purple}{\textbf{critical}}\\
  7 & & change to antonym & \textcolor{purple}{\textbf{critical}}\\
  8 & & change to negation & \textcolor{purple}{\textbf{critical}}\\
  9 & & replaced named entity & \textcolor{purple}{\textbf{critical}}\\
  10 & & incorrect numeric & \textcolor{purple}{\textbf{critical}}\\
  11 & & incorrect gender pronoun & \textcolor{purple}{\textbf{critical}}\\
 \midrule
  12 & \multirow{9}{*}{\rotatebox{90}{\textbf{fluency}}} & omitted conjunction & \textcolor{teal}{\textbf{minor}}\\
  13 & & part of speech shift & \textcolor{teal}{\textbf{minor}}\\
  14 & & switched word order (word swap) & \textcolor{teal}{\textbf{minor}}\\
  15 & & incorrect case (pronouns) & \textcolor{teal}{\textbf{minor}}\\
  16 & & incorrect preposition or article & \textcolor{teal}{\textbf{minor}}\textbf{-}\textcolor{violet}{\textbf{major}}\\
  17 & & incorrect tense & \textcolor{violet}{\textbf{major}}\\
  18 & & incorrect aspect & \textcolor{violet}{\textbf{major}}\\
  19 & & change to interrogative & \textcolor{violet}{\textbf{major}}\\
 \midrule
  20 & \multirow{5}{*}{\rotatebox{90}{\textbf{mixed}}} & omitted adj/adv & \textcolor{teal}{\textbf{minor}}\textbf{-}\textcolor{violet}{\textbf{major}}\\
  21 & & omitted content verb & \textcolor{purple}{\textbf{critical}}\\
  22 & & omitted noun & \textcolor{purple}{\textbf{critical}}\\
  23 & & omitted subject & \textcolor{purple}{\textbf{critical}}\\
  24 & & omitted named entity & \textcolor{purple}{\textbf{critical}}\\
  \midrule
  25 & \multirow{7}{*}{\rotatebox{90}{\textbf{typography}}} & misspelled word & \textcolor{teal}{\textbf{minor}}\\
  26 & & deleted character & \textcolor{teal}{\textbf{minor}}\\
  27 & & omitted final punctuation & \textcolor{teal}{\textbf{minor}}\\
  28 & & added punctuation & \textcolor{teal}{\textbf{minor}}\\
  29 & & tokenized sentence & \textcolor{teal}{\textbf{minor}}\\
  30 & & lowercased sentence & \textcolor{teal}{\textbf{minor}}\\
  31 & & first word lowercased & \textcolor{teal}{\textbf{minor}}\\
 \midrule
  32 & \multirow{3}{*}{\rotatebox{90}{\textbf{baseline}}} & empty string & \textcolor{orange}{\textbf{base}}\\
  33 & & unrelated translation & \textcolor{orange}{\textbf{base}}\\
  34 & & shuffled words & \textcolor{orange}{\textbf{base}} \\
  35 & & reference as translation & \textcolor{orange}{\textbf{base}}\\
\bottomrule
\end{tabular}}
\end{center}
\caption{List of perturbations included in \name\ with their corresponding error severity. Details can be found in Appendix \ref{sec:appendix}}
\label{tab:demetr_table}
\end{table}

\paragraph{Data sources and filtering:}
We utilize \emph{X-to-English} translation pairs from two different datasets, \textsc{WMT} \citep{callison-burch-etal-2009-findings, bojar-etal-2013-findings, bojar-etal-2015-findings, bojar-etal-2014-findings,  akhbardeh-etal-2021-findings,barrault-etal-2020-findings} and \textsc{FLORES} \citep{guzman-etal-2019-flores}, aiming at a wide coverage of topics from different sources. \textsc{WMT} has been widely used over the years as a popular MT shared task, while \textsc{FLORES} was recently curated to aid MT evaluation. We consider only the test split of each dataset to prevent possible leaks, as both  current and future metrics are likely to be trained on these two datasets. 
We sample \emph{100} sentences (\emph{50} from each of the two datasets) for each of the following \emph{10} languages:
French (\emph{fr}), Italian (\emph{it}), Spanish (\emph{es}), German (\emph{de}), Czech (\emph{cs}), Polish (\emph{pl}), Russian (\emph{ru}), Hindi (\emph{hi}), Chinese (\emph{zh}), and Japanese (\emph{ja}).\footnote{We choose languages that represent different families (Romance, Germanic, Slavic, Indo-Iranian, Sino-Tibetan, and Japonic) with different morphological traits (fusional, agglutinative, and analytic) and wide range of writing systems (Latin alphabet, Cyrillic alphabet, Devanagari script, Hanzi, and Kanji/Hiragana/Katakana).} We pay special attention to the language selection, as newer MT evaluation metrics, such as \textsc{Comet-QE} or \textsc{Prism-QE}, employ only the source text and the candidate translation.
We control for sentence length by including only sentences between \emph{15} and \emph{25} words long, measured by the length of the tokenized reference translation. Since we re-use the same sentences across multiple perturbations, we did not include shorter sentences because they are less likely to contain multiple linguistic phenomena of interest.\footnote{Similarly, we do not include sentences over 25 words long in \name\ as some languages may naturally allow longer sentences than others, and we wanted to control the length distribution.}  As the quality of sampled sentences varies, we manually check each source sentence and its translation to make sure they are of satisfactory quality.\footnote{In the sentences sampled from \textsc{WMT}, we notice multiple translation and grammar errors, such as translating Japanese \begin{CJK}{UTF8}{min}
その最大は本州列島で、世界で7番目に大きい島とされています。\end{CJK} as \textit{(the biggest being Honshu), making Japan the 7th largest island in the world}, which would suggest that Japan is an island, instead of \textit{the largest of which is the Honshu island, considered to be the seventh largest island in the world.} or "kakao" ("cacao") incorrectly declined as "kakaa" in Polish. These sentences were rejected, and new ones were sampled in their place. We also resampled sentences which translations contained artifacts from neighboring sentences due to partial splits and merges, and sentences which exhibit \textit{translationese}, that is sentences with source artifacts \citep{translationese}. Finally, we omit or edit sentences with translation artifacts due to the direction of translation. Both \textsc{WMT} and \textsc{FLORES} contain sentences translated from English to another languages. Since the translation process is \textit{not} always fully reversible, we omit sentences where translation from the give language to English would not be possible in the form included in these datasets (e.g., due to addition or omission of information).} 

\paragraph{Translating the data:}
Given the filtered collection of source sentences, we next translate them into English using the Google Translate API.\footnote{All sentences were translated in May, 2022.} We manually verify each translation, editing or resampling the instances where the machine translation contains critical errors. Through this process, we obtain \emph{1}K curated examples per perturbation (\emph{100} sentences $\times$ \emph{10} languages) that each consist of source and reference sentences along with a machine translation of reasonable quality.

\subsection{Perturbations in \name}
We perturb the machine translations obtained above in order to create \textit{minimal pairs}, which allow us to investigate the sensitivity of MT evaluation metrics to different types of errors. 
Our perturbations are loosely based on the Multidimensional Quality Metrics \citep[][MQM]{burchardt-2013-multidimensional} framework developed to identify and categorize MT errors. Most perturbations were performed semi-automatically by utilizing \textsc{Stanza} \citep{qi2020stanza-stanza}, \textsc{spaCy}\footnote{\url{https://spacy.io/usage/linguistic-features}} or \textsc{GPT-3}~\citep{brown2020language}, applying hand-crafted rules and then manually correcting any errors. Some of the more elaborate perturbations (e.g., translation by a too general term, where one had to be sure that a better, more precise term exists) were performed manually by the authors or linguistically-savvy freelancers hired on the Upwork platform.\footnote{See \url{https://www.upwork.com/}. Freelancers were paid an equivalent of \$15 per hour.} Special care was given to the plausibility of perturbations (e.g., numbers for replacement were selected from a probable range, such as \emph{1}-\emph{12} for months). See Table~\ref{tab:demetr_table_details_small} for descriptions and examples of most perturbations; full list in Appendix \ref{sec:appendix}.

We roughly categorize our perturbations into the following four categories:

\begin{table*}
\small
\begin{center}
\renewcommand*{\arraystretch}{1.2}
\resizebox{\textwidth}{!}{\begin{tabular}{cp{2cm}p{11cm}p{9cm}p{2cm}c} 
 \toprule
  \bf Category & \bf \makecell{Type} & \bf Example & \bf Description & \bf Implementation & \bf Error Severity \\
 \midrule

   \multirow{32}{*}{\rotatebox{90}{\Large{\textbf{\color{purple}ACCURACY}}}} & \textbf{\makecell{repetition}} & I don't know if you realize that most of the goods imported into this country from Central America are duty \textbf{\color{purple}free}.
   
I don't know if you realize that most of the goods imported into this country from Central America are duty \textbf{\color{purple}free free}. & The last word is being repeated twice. Punctuation is added after the last repeated word. & automatic & minor\\
   & \textbf{\makecell{repetition}} & Gordon Johndroe, Bush's spokesman, referred to the North Korean commitment as "an important advance towards the goal of achieving verifiable denuclearization of the Korean \textbf{\color{purple}penisula}." 
   
  Gordon Johndroe, Bush's spokesman, referred to the North Korean commitment as "an important advance towards the goal of achieving verifiable denuclearization of the Korean \textbf{\color{purple}penisula penisula penisula penisula}." & The last word is being repeated four times. Punctuation is added after the last repeated word. & automatic & minor\\
   & \textbf{\makecell{hypernym}} & The language most of the people working in the Vatican City use on a daily basis is Italian, and Latin is often used in religious \textbf{\color{purple}ceremonies.}
   
The language most of the people working in the Vatican City use on a daily basis is Italian, and Latin is often used in religious \textbf{\color{purple}activities.} & A word translated by a too general term (undertranslation). Special care was given in order to assure the word used in perturbed text is more general, and incorrect, translation of the original word. & manual with suggestions from \textsc{GPT-3} & major\\

   & \textbf{\makecell{untranslated}} & The Polish Air Force will eventually be equipped with 32 F-35 Lightning II fighters \textbf{\color{purple}manufactured} by Lockheed Martin. 
   
The Polish Air Force will eventually be equipped with 32 F-35 Lightning II fighters \textbf{\color{purple}produkowane} by Lockheed Martin. & One word is being left untranslated. We manually assure that each time only one word is left untranslated. & manual & major\\

   & \textbf{\makecell{completeness}} & She is \textbf{\color{purple}in custody} pending prosecution and trial; but any witness evidence could be negatively impacted because her image has been widely published.

She is \textbf{\color{purple}\_\_\_\_\_} pending prosecution and trial; but any witness evidence could be negatively impacted because her image has been widely published. & One prepositional phrase is being removed. Whenever possible, we remove the shortest prepositional phrase in order to assure that the perturbed sentence is not much shorter than the original translation. & automatic (Stanza) with manual check & major\\
   & \textbf{\makecell{addition}} & \textbf{\color{purple}\_\_\_\_\_} Plants look their best when they are in a natural environment, so resist the temptation to remove "just one."

 \textbf{\color{purple}Power} plants look their best when they are in a natural environment, so resist the temptation to remove "just one." & One word is being added. We make sure that the added word does not disturb the grammaticality of the sentence but changes the meaning in a significant way. & manual & critical\\
   & \textbf{\makecell{antonym}} & He has been unable to relieve the \textbf{\color{purple}pain} with medication, which the competition prohibits competitors from taking.

He has been unable to relieve the \textbf{\color{purple}pleasure} with medication, which the competition prohibits competitors from taking. & One word (noun, verb, adj., or adv.) is being changed to its antonym. & manual with suggestions from \textsc{GPT-3} & critical\\
   & \textbf{\makecell{mistranslation \\ negation}} & Last month, a presidential committee \textbf{\color{purple}recommended} the resignation of the former CEP as part of measures to push the country toward new elections.

Last month, a presidential committee \textbf{\color{purple}didn't recommend} the resignation of the former CEP as part of measures to push the country toward new elections.  & Affirmative sentences are being changed into negations. Rare negations are being changed to affirmative sentences. & manual & critical\\
   & \textbf{\makecell{mistranslation \\ named entity}} & Late night presenter \textbf{\color{purple}Stephen Colbert} welcomed 17-year-old Thunberg to his show on Tuesday and conducted a lengthy interview with the Swede.

Late night presenter \textbf{\color{purple}John Oliver} welcomed 17-year-old Thunberg to his show on Tuesday and conducted a lengthy interview with the Swede. & Named entity is replaced with another named entity from the same category (person, geographic location, and organization). & automatic (Stanza) with manual check & critical\\
   & \textbf{\makecell{mistranslation \\ numbers}} & The Chinese Consulate General in Houston was established in \textbf{\color{purple}1979} and is the first Chinese consulate in the United States.

The Chinese Consulate General in Houston was established in \textbf{\color{purple}1997} and is the first Chinese consulate in the United States. & A number is being replaced with an incorrect one. Special attention was given to keep the numerals with resonable/common range for the given category (e.g., 0-100 for percentages; 1-12 for months). We also assure that the replacement will not create an illogical sentence (e.g., replacing ``1920'' with ``1940'' in ``from 1920 to 1930'') & manual & critical\\
   & \textbf{\makecell{mistranslation \\ gender}} & \textbf{\color{purple}He} has been unable to relieve the pain with medication, which the competition prohibits competitors from taking.

\textbf{\color{purple}She} has been unable to relieve the pain with medication, which the competition prohibits competitors from taking. & Exactly one feminine pronoun in the sentence (such as ``she'' or ``her'') is being with a masculine pronouns (such as ``he'' or ``him'') or vice-versa. This includes reflexive pronouns (i.e., ``him/herself'') and possessive adjectives (i.e., ``his/her''). & automatic with manual check & critical\\
 \midrule
   \multirow{23}{*}{\rotatebox{90}{\Large{\textbf{\color{purple}FLUENCY}}}} & \textbf{\makecell{cohesion}}  & Scientists want to understand how planets have formed \textbf{\color{purple}since} a comet collided with Earth long ago, and especially how Earth has formed.

Scientists want to understand how planets have formed \textbf{\color{purple}\_\_\_\_\_} a comet collided with Earth long ago, and especially how Earth has formed. & A conjunction, such as ``thus'' or ``therefore'' is removed. Special attention was given to keep the rest of the sentence unperturbed. & automatic (spaCy) with manual check & minor\\
   & \textbf{\makecell{grammar \\ pos shift }}& The U.S. Supreme Court last year blocked the Trump \textbf{\color{purple}administration} from including the citizenship question on the 2020 census form.

The U.S. Supreme Court last year blocked the Trump \textbf{\color{purple}administrate} from including the citizenship question on the 2020 census form. & Affix of the word is being changed keeping the stem kept constant (e.g., ``bad'' to ``badly'') which results in the part-of-speech shift. The degree to which the original meaning is affected varies, however, the intended meaning is easily retrivable from the stem and context. & manual & minor\\
   & \textbf{\makecell{grammar \\ swap order}} & I don't know if you realize that most of the goods imported \textbf{\color{purple}into this} country from Central America are duty free.

I don't know if you realize that most of the goods imported \textbf{\color{purple}this into} country from Central America are duty free. & Two neighboring words are being swapped to mimic word order error. & automatic (spaCy) & minor\\
   & \textbf{\makecell{grammar \\ case}} & \textbf{\color{purple}She} announced that after a break of several years, a Rakoczy horse show will take place again in 2021.

\textbf{\color{purple}Her} announced that after a break of several years, a Rakoczy horse show will take place again in 2021. & One pronoun in the sentence is being changed into a different, incorrect, case (e.g., ``he'' to ``him''). & automatic (spaCy) with manual check & minor\\
   & \textbf{\makecell{grammar \\ function word}} & Last month, \textbf{\color{purple}a} presidential committee recommended the resignation of the former CEP as part of measures to push the country toward new elections.

Last month, \textbf{\color{purple}an} presidential committee recommended the resignation of the former CEP as part of measures to push the country toward new elections. & A preposition or article is being changed into an incorrect one to mimic mistake in function words usage. While most perturbations result in minor mistakes (i.e., the original meaning is easily retrivable) some may be more severe. & automatic with manual check & minor-major\\
   & \textbf{\makecell{grammar \\ tense}} & Cyanuric acid and melamine \textbf{\color{purple}were} both found in urine samples of pets who died after eating contaminated pet food.

Cyanuric acid and melamine \textbf{\color{purple}are} both found in urine samples of pets who died after eating contaminated pet food. & A tense is being change into an incorrect one. We consider past, present, as well as the future tense (although this may be classified as modal verb in English) & manual & major\\
   & \textbf{\makecell{grammar \\ aspect}} & He \textbf{\color{purple}has been} unable to relieve the pain with medication, which the competition prohibits competitors from taking.

He \textbf{\color{purple}is being} unable to relieve the pain with medication, which the competition prohibits competitors from taking. & Aspect is being changed to an incorrect one (e.g., perfective to progressive) \textit{without} changing the tense. & manual & major\\
   & \textbf{\makecell{grammar \\interrogative}} & \textbf{\color{purple}This is} the tenth time since the start of the pandemic that Florida's daily death toll has surpassed 100.

\textbf{\color{purple}Is this} the tenth time since the start of the pandemic that Florida's daily death toll has surpassed 100? & Affirmative mood is being changed to interrogative mood. & manual & major\\
 \midrule
   \multirow{15}{*}{\rotatebox{90}{\Large{\textbf{\color{purple}MIXED}}}} & \textbf{\makecell{omission \\ adj/adv}} & Rangers \textbf{\color{purple}closely} monitor shooters participating in supplemental pest control trials as the trials are monitored and their effectiveness assessed.

Rangers \textbf{\color{purple}\_\_\_\_\_} monitor shooters participating in supplemental pest control trials as the trials are monitored and their effectiveness assessed. & An adjective or adverb is being removed. While in most cases this leads to & automatic (spaCy) with manual check & minor-major\\
   & \textbf{\makecell{omission \\ content verb}} & Catri \textbf{\color{purple}said} that 85\% of new coronavirus cases in Belgium last week were under the age of 60.

Catri \textbf{\color{purple}\_\_\_\_\_} that 85\% of new coronavirus cases in Belgium last week were under the age of 60. & Content verb is being removed (this excludes auxilary verbs and copulae). & Automatic with manual check & critical\\
   & \textbf{\makecell{omission \\ noun}} & In 1940 he stood up to other government \textbf{\color{purple}aristocrats} who wanted to discuss an "agreement" with the Nazis and he very ably won.

In 1940 he stood up to other government \textbf{\color{purple}\_\_\_\_\_} who wanted to discuss an "agreement" with the Nazis and he very ably won.

 & Noun, which is not a named entity or a subject, is being removed. We remove the head of the noun phrase including compound nouns. & automatic (spaCy) with manual check & critical\\
   & \textbf{\makecell{omission \\ subject}}& His \textbf{\color{purple}research} shows that the administration of hormones can accelerate the maturation of the baby's fetal lungs.

His \textbf{\color{purple}\_\_\_\_\_} shows that the administration of hormones can accelerate the maturation of the baby's fetal lungs.

 & Subject is being removed. We remove the head of the noun phrase including compound nouns. & automatic (spaCy) with manual check & critical\\
   & \textbf{\makecell{omission \\ named entry}}& I don't know if you realize that most of the goods imported into this country from \textbf{\color{purple}Central America} are duty free.

I don't know if you realize that most of the goods imported into this country from \textbf{\color{purple}\_\_\_\_\_} are duty free.
 & Named entity, which is not a subject, is being removed. & automatic (Stanza) with manual check & critical\\
 
\bottomrule
\end{tabular}}
\end{center}
\caption{A subset of perturbations in \name~along with examples (detailed changes are highlighted in {\color{purple} purple}). A full list of perturbations is provided in ~\autoref{tab:demetr_table_details1} and ~\autoref{tab:demetr_table_details2} in Appendix \ref{sec:appendix}.}
\label{tab:demetr_table_details_small}
\end{table*}

\begin{itemize}

    \item \textsc{\textbf{Accuracy}:} Perturbations in the accuracy category modify the semantics of the translation by either incorporating misleading information (e.g., by adding plausible yet inadequate text or changing a word to its antonym) or omitting information (e.g., by leaving a word untranslated).

    \item \textsc{\textbf{Fluency}:} Perturbations in the fluency category focus on grammatical accuracy (e.g., word form agreement, tense, or aspect) and on overall cohesion. Compared to the mistakes in the accuracy category, the true meaning of the sentence can be usually recovered from the context more easily.

    \item \textsc{\textbf{Mixed}:} Certain perturbations can be classified as both   accuracy and fluency errors. 
    Concretely, this category consists of omission errors that not only obscure the meaning but also affect the grammaticality of the sentence. One such error is \emph{subject removal}, which will result not only in an ungrammatical sentence, leaving a gap where the subject should come, but also in information loss.
    
    \item \textsc{\textbf{Typography}:} This category concerns punctuation and minor orthographic errors. Examples of mistakes in this category include punctuation removal, tokenization, lowercasing, and common spelling mistakes.
    
    \item \textsc{\textbf{Baseline}:} Finally, we include both upper and lower bounds, since learned metrics such as \textsc{Bleurt} and \textsc{Comet} do not have a specified range that their scores can fall into. Specifically, we provide three baselines: as lower bounds, we either change the
    translation to an unrelated one or provide an empty string,\footnote{Since most of the metrics will not accept an empty string, we pass a full stop instead.} while as an upper bound, we set the perturbed translation $t'$ equal to the reference translation $r$, which should return the highest possible score for reference-based metrics. 

\end{itemize}

\paragraph{Error severity:}
Our perturbations can also be categorized by their \emph{severity} (see Table \ref{tab:demetr_table}). We use the following categorization scheme for our analysis experiments:

\begin{itemize}

    \item \textsc{\textcolor{teal}{\textbf{MINOR}}:} In this type of error, which includes perturbations such as dropping punctuation or using the wrong article,  the meaning of the source sentence can be easily and correctly interpreted by human readers. 
    
    \item \textsc{\textcolor{violet}{\textbf{MAJOR}}:} Errors in this category may not affect the overall fluency of the sentence but will result in some missing details. Examples of major errors include undertranslation (e.g., translating ``church'' as ``building''), or leaving a word in the source language untranslated.  
    
    \item \textsc{\textcolor{purple}{\textbf{CRITICAL}}:} These are catastrophic errors that result in crucial pieces of information going missing or incorrect information being added in a way unrecognizable for the reader, and are also likely to suffer from severe fluency issues. Errors in this category include subject deletion or replacement of a named entity.
    
\end{itemize}

\section{Performance of MT evaluation metrics on \name}
\label{sec:experiments}

We test the accuracy and sensitivity of \emph{14} popular MT evaluation metrics on the perturbations in \name. We include both traditional string-based metrics, such as \textsc{Bleu} or \textsc{ChrF}, as well as newer learned metrics, such as \textsc{Bleurt} and \textsc{Comet}. Within the latter category, we also include two reference-free metrics, which rely only on the source sentence and translation and open possibilities for a more robust MT evaluation. The rest of this section provides an overview of the evaluation metrics before analyzing our findings. Detailed results of each metric on every perturbation are found in Table~\ref{tab:significance5}.

\subsection{Evaluation metrics}

String-based metrics can be used to evaluate \textit{any} language, provided the availability of a reference translation (see Table \ref{tab:metrics_table}). Their score is a function of string overlap or edit-distance, though it may not be always easily interpretable \citep{muller_2020}. Only \textsc{Bleu}\footnote{For all string-based metrics we use the HuggingFace implementations available at \url{https://huggingface.co/evaluate-metric}. In the case of \textsc{Bleu}, we use the SacreBLEU version 2.1.0~\cite{post-2018-call}.} allows for multiple references in order to account for many possible translations of a sentence; however, it is rarely used with more than one reference due to the lack of multi-reference datasets \citep{mathur-etal-2020-tangled}. 
Learned metrics, on the other hand, are much less transparent. \textsc{BERTScore} relies on contextualized embeddings, while \textsc{Prism} employs zero-shot paraphrasing. \textsc{Comet} and \textsc{Bleurt} directly fine-tune pretrained language models on human judgments provided as Direct Assessments or MQM annotations.\footnote{We use the HuggingFace implementation of \textsc{BERTScore}, \textsc{Bleurt-20}, \textsc{Comet}, and \textsc{Comet-QE}. For \textsc{Bleurt-20}, we use \textsc{Bleurt-20}, the most recent and recommended checkpoints, for \textsc{Comet} and \textsc{Comet-QE} we use the SOTA models from \textsc{WMT21} shared task, wmt21-comet-mqm and wmt21-comet-qe-mqm checkpoints, and for \textssc{BERTScore} we use roberta-large. For PRISM, we use the implementation available at~\url{https://github.com/thompsonb/prism}}

\begin{table}[t!]
\small
\begin{center}
\resizebox{\columnwidth}{!}{\begin{tabular}{lrr} 
 \toprule
 Metric & \# Params & Language \\
 \midrule
 \multicolumn{3}{c}{\textit{string-based metrics}}\vspace{0.1cm}\\
  \textsc{Bleu} \citep{papineni-etal-2002-bleu} & -- & any\\
  \textsc{CER} \citep{morris2004} & --  & any\\
  \textsc{chrF} \citep{popovic-2015-chrf} & --  & any \\
  \textsc{chrF2} \citep{popovic-2017-chrf}        & --                & any\\
  \textsc{Meteor} \citep{banerjee-lavie-2005-meteor} & -- &  any\\
  \textsc{Rouge-2} \citep{lin-2004-rouge}  &  --  & any\\
  \textsc{TER} \citep{snover-etal-2006-study} & -- & any \\
  \midrule
    \multicolumn{3}{c}{\textit{pre-trained metrics}}\vspace{0.1cm}\\
  \textsc{BARTScore} \citep{NEURIPS2021-bartscore} &  406M  & 50\\
  \textsc{BERTScore} \citep{bert-score} &  355M  & 104\\
  \textsc{BleurtT-20} \citep{bleurt}  &  579M  & 104\\
  \textsc{Comet} \citep{rei-etal-2021-references} & 580M  & 100\\
  \textsc{Prism} \citep{thompson-post-2020-automatic} & 745M  & 39\\
  \midrule
    \multicolumn{3}{c}{\textit{pre-trained reference-free metrics}}\vspace{0.1cm}\\
  \textsc{Comet-QE} \citep{rei-etal-2021-references} & 569M & 100\\
  \textsc{Prism-QE} \citep{thompson-post-2020-automatic} & 745M & 39\\
\bottomrule
\end{tabular}}
\end{center}
\caption{Details of metrics tested on \name. We report the parameter count for the largest available checkpoint of each learned metric. For learned metrics, we report the maximum number of languages that each can accept as input. While most of the learned metrics leverage pretrained multilingual language models (e.g., mBERT), it is important to note that they have not been validated against human judgments of MT quality on all of these languages (e.g., BLEURT-20 is only validated on 13 languages).}
\label{tab:metrics_table}
\end{table}

\begin{table}[t!]
\small
\begin{center}
\begin{tabular}{ lrrrrr } 
 \toprule
 Metric & Base & Crit. & Maj. & Min. & All \\
 \midrule
 \multicolumn{6}{c}{\textit{string-based metrics}}\vspace{0.1cm}\\
   \textsc{Bleu} & 100.00&	80.29&	83.43&	72.49&	78.70\\
  \textsc{CER} & 99.15&	80.37&	83.59&	80.20&	81.88\\
   \textsc{chrF} & 100.00&	91.13&	90.89&	81.23&	87.54\\
      \textsc{chrF2} & 100.00&	91.27&	92.21&	83.68&	\textbf{88.80}\\
     \textsc{Meteor} & 100.00&	82.95&	79.69&	58.97&	73.60\\
      \textsc{Rouge-2} & 99.90&	76.91&	80.99&	47.10&	66.58\\
   \textsc{TER} & 99.20&	72.57&	77.93&	59.13&	69.39\\
   \midrule
    \multicolumn{6}{c}{\textit{learned metrics}}\vspace{0.1cm}\\
    \textsc{BARTScore} & 100.00 & 95.11 & 89.68 & 79.48 & 88.16 \\
    \textsc{BERTScore} & 100.00 & 98.11 &	96.22 &	98.50 &	98.11 \\
  \textsc{Bleurt-20} & 100.00&	98.78&	95.63&	97.98&	98.06 \\
   \textsc{Comet} & 100.00&	96.24&	92.96&	93.46&	94.83\\
   \textsc{Prism} & 100.00&	98.74&	97.51&	99.44&	\textbf{98.92}\\
  \textsc{Comet-QE} & 77.80&	84.49&	76.73&	89.85&	85.16\\
   \textsc{Prism-QE} & 97.40&	96.70&	95.68&	99.21&	97.63\\
\bottomrule
\end{tabular}
\end{center}
\caption{Accuracy on \name\ perturbations for both string-based and learned metrics, shown bucketed by error severity (baseline, critical, major, and minor errors) as well as averaged across all perturbations. Baseline accuracies were computed excluding the \emph{reference as translation} identity perturbation. Detailed accuracies for all perturbations along with the significance testing are shown in \autoref{tab:significance5} in the Appendix \ref{sec:appendix}.
}
\label{tab:demeter_accuracy}
\end{table}

\subsection{Perturbation accuracy}
First, we measure the \emph{accuracy} of each metric on \name. For each perturbation, we define the accuracy as the percentage of the time that \texttt{SCORE}$(r, t)$ is greater than \texttt{SCORE}$(r, t')$.\footnote{We do not give metrics credit for giving an equal score to both perturbed and unperturbed sentences.} Since all perturbed sentences are \textit{less correct} versions of the original machine translation, we expect all metrics to perform well on this task. Table \ref{tab:demeter_accuracy} contains the accuracies averaged across both error severity as well as overall. Interesting results include:

\paragraph{Learned metrics achieve higher accuracy than string-based ones:}
All but two learned metrics (\textsc{BARTScore} and \textsc{Comet-QE}) achieve around or over \emph{95\%} accuracy,\footnote{This is true even for \textsc{Prism-QE}, whose base neural MT model does not support Hindi but still manages to perform decently without the source.} which is to be expected, as each perturbation \textit{clearly} affects the quality of the translation, though to varying degrees. \textsc{Prism} is the most accurate metric on \name, reaching an accuracy of \emph{98.92\%}. 
Performance of string-based metrics, on the other hand, is alarmingly bad. \textsc{Bleu}, often the \textit{only} metric employed to evaluate the MT output \citep{marie-etal-2021-scientific}, achieves an overall accuracy of only \emph{78.70\%}. To illustrate their struggles, the accuracy of string-based metrics ranges from \emph{54\%} to \emph{84\%} on the adjective/adverb removal perturbation, where a single adjective or adverb is omitted.

The best performing string-based metric is \textsc{chrF2}, which corroborates results reported in \citet{shipornotKomci}.

\paragraph{\textsc{Prism-QE} achieves better accuracy than \textsc{Comet-QE} for reference-free metrics:} 
Of the two reference-free metrics we evaluate, we notice that \textsc{Comet-QE} struggles with some perturbations. Most notably, its accuracy when given a \emph{random} translation (i.e., a translation that does not match the source sentence) oscillates around \emph{50\%} (chance level) across all languages. Furthermore, \textsc{Comet-QE} shows low accuracy on gender (i.e., masculine pronouns replaced with feminine pronouns or vice-versa), number (i.e., a number replaced for another, reasonable number), and interrogatives (i.e., change of affirmative mood into interrogative mood). \textsc{Comet-QE} also strongly prefers (\emph{88\%}) the translation stripped of final punctuation over the complete sentence, in comparison to \emph{0\%} for \textsc{Prism-QE}. In terms of accuracy, \textsc{Prism-QE} performs exceptionally well on all perturbations, achieving lower accuracies (yet still around \emph{80\%}) only for Hindi---a language it was not trained on.

\section{Sensitivity analysis}
\label{sec:analysis}

While the accuracy of a metric on \name\ is useful to know, it also obscures the \emph{sensitivity} of a metric to a particular perturbation. Are metrics more sensitive to \textsc{critical} errors than \textsc{minor} ones? Are different \textit{learned} metrics comparatively more or less sensitive to a particular perturbation?  In this section, we explore these questions and highlight interesting observations, focusing primarily on the behavior of learned metrics.

\paragraph{Measuring sensitivity:}
Since each of our metrics has a different score range, we cannot na\"ively just compare their score differences to analyze sensitivity. Instead, we compute a ratio that intuitively answers the following question: how much does \texttt{SCORE} drop on this perturbation compared to the catastrophic error of producing an empty string? We choose the empty string as a control since it is the perturbation that results in the largest \texttt{SCORE} drop for most metrics. Concretely, for a given reference translation $r_i$, machine translation $t_i$, and perturbed translation $t'_i$, we compute a ratio $z_i$ as:


\begin{figure*}[ht!]
    \centering
    \includegraphics[width=0.9\textwidth]{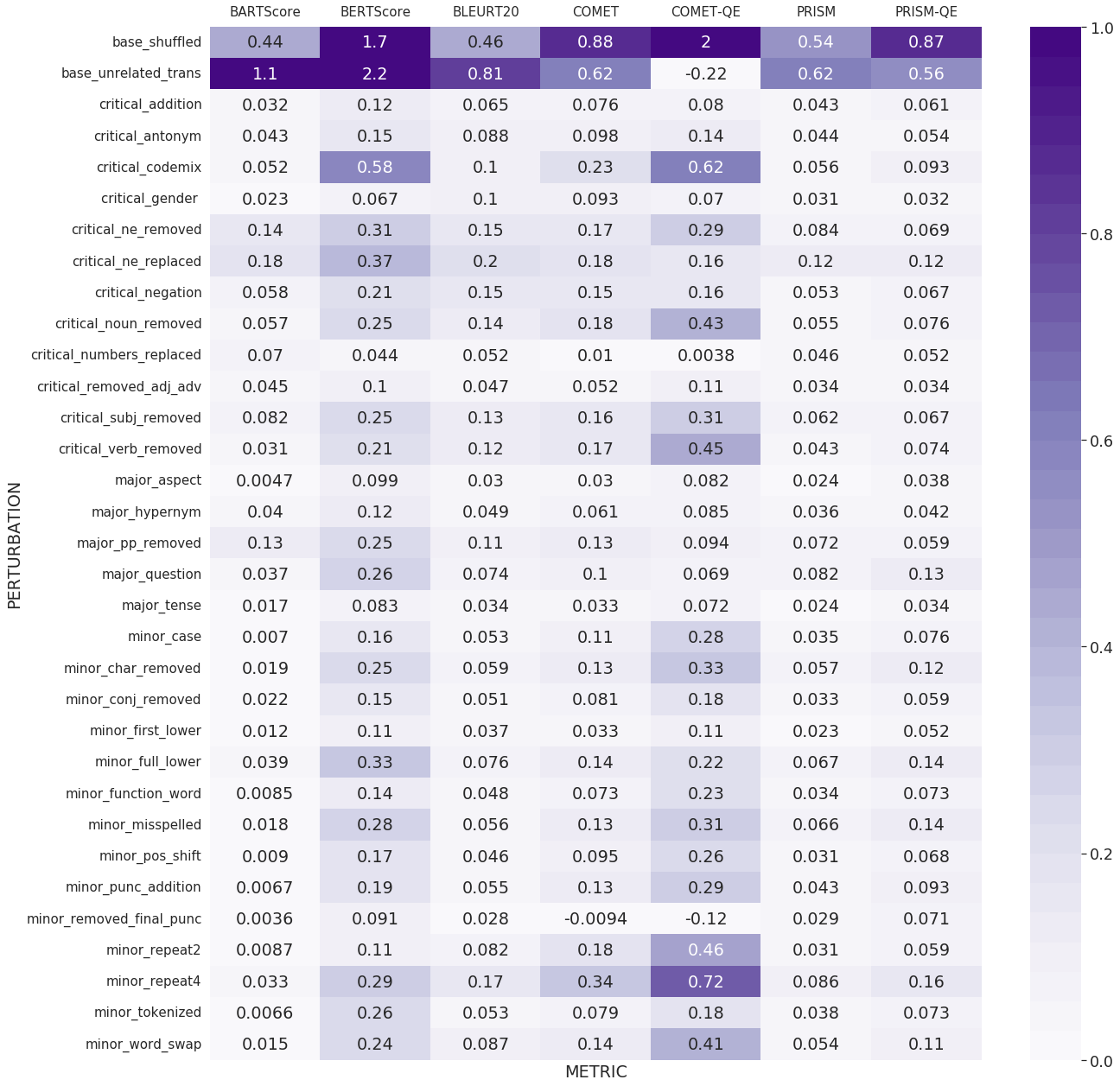}
    \caption{A heatmap of the sensitivity of learned metrics to different perturbations in \name. The numbers are the ratios $z$ computed as described in Section \ref{sec:analysis}. Higher values denote higher relative sensitivity to the perturbation and are marked by a darker color. The error severity categories are arranged from \textit{\textbf{minor}} (bottom part) through \textit{\textbf{major}} (middle part) to \textit{\textbf{critical}} (upper part). The last two errors are baselines.}
    \label{fig:galaxy}
\end{figure*}

\begin{equation}
    z_i = \frac{\texttt{SCORE}(r_i, t_i) - \texttt{SCORE}(r_i, t'_i)}{\texttt{SCORE}(r_i, t_i) - \texttt{SCORE}(r_i, \emph{empty string})}
\end{equation}

Then, for each perturbation category, we aggregate the example-level ratios to obtain $z$ by simply taking a mean, $z = \sum_i \frac{z_i}{N}$,
where N is the number of examples for that perturbation (in most cases, \emph{1}K).\footnote{The ratio is a reasonable but also a rough estimate of metric sensitivity. Since it depends highly on the scores given by the metric to an empty string, we also make sure that all tested metrics achieve an accuracy close to 100\% and can significantly distinguish between an empty string and the actual translation.} Figure~\ref{fig:galaxy} contains a heatmap plotting this $z$ ratio for each perturbation and learned metric, and forms the core of the following analysis. 

\paragraph{\textsc{BERTScore} is relatively more sensitive to some minor errors than it is to critical errors:} Although we observe that \textsc{BERTScore} drops only by a small absolute number for most perturbations, it is actually quite sensitive to many perturbations, especially when passing an unrelated translation and a shuffled version of the existing translation -- two of the most drastic perturbations. It also shows higher sensitivity to untranslated words (i.e., codemixing) than to the remaining perturbations, which is to be expected as \textsc{BERTScore} uses a multilingual model. However, its sensitivity to incorrect numbers (\emph{0.044}), gender information (\emph{0.067}), or aspect change (\emph{0.099}) is lower than sensitivity to less severe errors, such as tokenized sentence (\emph{0.26}) or lower-cased sentence (\emph{0.33}) -- a trend visible in other metrics, though not to such an extent.

\paragraph{\textsc{COMET-QE}, a metric adapted to MQM scoring, does not perform well on \name:} \textsc{Comet-QE} trained on MQM ratings (i.e., on the identification of mistakes similar to those included in \name) varies in its sensitivity to perturbations. While it is sensitive to a sentence with shuffled words, it is not sensitive to a different, unrelated translation (an observation in line with its accuracy). \textsc{Comet-QE} also seems to be insensitive to minor errors such as the removal of the final punctuation, but also to some major or critical errors such as gender and number replacement.\footnote{Welsch \textit{t}-test also reveals that the difference between the scores for the original MT and perturbed text is not significant (\textit{p}-val>\emph{.05})} Furthermore, \textsc{Comet-QE} is much more sensitive to word repetition (\emph{0.46}-\emph{0.72}) and word swap (\emph{0.41}) than to some critical or major errors, such as  named entity replacement (\emph{0.16}) or sentence negation (\emph{0.16}). Overall, \textsc{Comet-QE} behaves very differently from most of the other metrics, and in ways that are difficult to explain.

\paragraph{Overall, all metrics struggle to differentiate between minor and critical errors:} While all metrics other than \textsc{Comet-QE} are very sensitive to the two baselines (different translation and shuffled words) when compared to other perturbations (\emph{0.44}-\emph{2.20}), they struggle to differentiate the severity of some critical errors, such as an addition of a plausible but meaning-changing word (\emph{0.032}-\emph{0.12}) or incorrect number (\emph{0.0038}-\emph{0.07}). These ratios are lower than of some minor errors such as a word repeated four times (\emph{0.086}-\emph{0.72}). In fact, \textsc{BERTScore}, \textsc{Comet}, and \textsc{Comet-QE} are \textit{more} sensitive to word repetition than to an addition of a word which ultimately critically changes the meaning.

\section{Related Work}
\label{sec:related_work}

Our work builds on the previous efforts to analyze the performance of MT evaluation metrics, as well as efforts to curate diagnostic datasets for NLP.

\paragraph{Analysis of MT evaluation metrics:}

\citet{fomicheve-et-al-2019} show that metric performance varies significantly across different levels of MT quality. \citet{freitag-etal-2020-bleu} demonstrate the importance of reference quality during evaluation. \citet{shipornotKomci} investigate the performance of pretrained and string-based metrics, and conclude that learned metrics outperform string-based metrics, with \textsc{Comet} being the best-performing metric at the time. However, \citet{comet-fails-num-ne-2022} explore \textsc{Comet} models in more depth finding, just as in the current study, that the models are \textit{not} sensitive to number and named entity errors. \citet{hanna-bojar-2021-fine}, on the other hand, find that \textsc{BERTScore} is more robust to errors in major content words, and less so to small errors. Finally, \citet{billbords} introduce a leaderboard for generation tasks that ensembles many of the metrics used here.

\paragraph{Diagnostic datasets:}

A number of previous studies employed diagnostic tests to explore the performance of NLP models.
\citet{marvin-linzen-2018-targeted} evaluate abilities of LSTM based language models to rate grammatical sentence higher than ungrammatical ones by curating a dataset of  minimal pairs in English. \citet{warstadt-etal-2020-blimp-benchmark} also utilize the concept of linguistic minimal pairs to evaluate the sensitivity of language models to various linguistic errors. \citet{ribeiro-etal-2020-beyond} curate a checklist of perturbations to test the robustness of general NLP models.

\citet{specia-etal-2010-dataset} introduce a simplified dataset of translations by four MT systems annotated for their quality in order to evaluate MT evaluation metrics.
\citet{sai2021perturbation} also propose a checklist-style method to test the robustness of evaluation metrics for MT; however, they limit themselves to Chinese-to-English translation. Furthermore, many of the perturbations introduced in \citet{sai2021perturbation} does not control for a single aspect, as \name\ does, and are not manually verified. \citet{macketanz-etal-2018-tq}, on the other hand, design a linguistic test suite to evaluate the quality of MT from German to English, which WMT21 \citep{wmt-2021-machine} utilizes as a challenge dataset for MT evaluation metrics. Finally, \citet{wmt-2021-machine} create a nine-category challenge set from a Chinese to English corpus, in order to test MT evaluation metrics, that are being submitted to the shared task.

\section{Conclusion}
\label{sec:conclusion}

We present \name, a dataset designed to diagnose MT evaluation metrics. \name\ consists of \emph{31}K semi-automatically generated perturbations that cover \emph{35} different linguistic phenomena. Our experiments showed that learned metrics are notably better than \textit{any} string-based metrics at distinguishing perturbed from unperturbed translations, which confirms results reported in other studies \citep{shipornotKomci, fomicheve-et-al-2019}. We further explore the sensitivity of learned metrics, showing that even the best-performing metrics struggle to distinguish between minor errors such as word repetition and critical errors such as incorrect number, aspect, and gender. We will publicly release \name\ to spur more informed future development of machine translation evaluation metrics. 

\section*{Limitations}
\label{sec:limitations}

While \name\  incorporates a wide range of linguistic phenomena, including various semantic, pragmatic, and morphological errors, all examples included in \name\ are of translations \textit{into}-English. It is likely that other translation directions may introduce other errors or metrics may be more/less sensitive to them. Furthermore, 
Furthermore, we decided to utilize sentence level translation as most metrics evaluate the translation on the sentence level and
to highlight specific errors, which could be less apparent in the paragraph level setup. 
However, sentence level data cannot model discourse level errors, which remain an open problem in both machine translation and its evaluation. Furthermore, as \name\ was constructed using \textsc{WMT} and \textsc{Flores} the domains incorporated in \name\ are restricted to the ones present in these two datasets (i.e., mostly news and informational materials).
Finally, even though in most cases multiple correct translations of the source sentence exist, we provide only one reference. We decided not to include multiple reference due to the time restrictions as well as the fact that the only metric currently supporting multiple references is \textsc{Bleu}.

\section*{Ethical Considerations}

Some perturbations were conducted manually with a help of freelancers hired on Upwork. The freelancers were informed of the purpose of this experiment. They were paid an equivalent of \$15 per hour. We also adjusted this hourly rate to cover the 20\% Upwork charge, which the platform charges the freelancers. 
\section*{Acknowledgements}

We would like to show our appreciation for the annotators hired on Upwork who took time and effort to fully understand the project and who helped us to assure good quality of the data. We would also like to thank Sergiusz Rzepkowski, Paula Kurzawska, and Kalpesh Krishna for their help in verifying the quality of translation and/or the perturbations. Finally, we would like to thank the reviewers for their constructive comments which helped to shape this paper, as well as UMass NLP community for their insights and discussions during this project. This project was partially supported by awards IIS-1955567 and IIS-2046248 from the National Science Foundation (NSF).


\bibliography{bib/anthology,bib/custom}
\bibliographystyle{bib/acl_natbib}

\clearpage
\setcounter{table}{0}
\renewcommand{\thetable}{A\arabic{table}}
\appendix
\section{Appendix}
\label{sec:appendix}

\begin{table*}[!ht]
\small
\begin{center}
\renewcommand*{\arraystretch}{1.5}
\resizebox{\textwidth}{!}{\begin{tabular}{ccp{2cm}p{10cm}p{7cm}p{2cm}c} 
 \toprule
 \bf {\large ID} & \bf {\large Category} & \bf {\large Type} & \bf {\large Example} & \bf {\large Description} & \bf {\large Application} & \bf  {\large Error Severity} \\
 \midrule

   1 & \multirow{32}{*}{\rotatebox{90}{\LARGE ACCURACY}} & repetition & I don't know if you realize that most of the goods imported into this country from Central America are duty \textbf{\color{red}free}.
   
I don't know if you realize that most of the goods imported into this country from Central America are duty \textbf{\color{red}free free}. & The last word is being repeated twice. Punctuation is added after the last repeated word. & automatic & minor\\
   2 & & repetition & Gordon Johndroe, Bush's spokesman, referred to the North Korean commitment as "an important advance towards the goal of achieving verifiable denuclearization of the Korean \textbf{\color{red}penisula}." 
   
  Gordon Johndroe, Bush's spokesman, referred to the North Korean commitment as "an important advance towards the goal of achieving verifiable denuclearization of the Korean \textbf{\color{red}penisula penisula penisula penisula}." & The last word is being repeated four times. Punctuation is added after the last repeated word. & automatic & minor\\
   3 & & hypernym & The language most of the people working in the Vatican City use on a daily basis is Italian, and Latin is often used in religious \textbf{\color{red}ceremonies.}
   
The language most of the people working in the Vatican City use on a daily basis is Italian, and Latin is often used in religious \textbf{\color{red}activities.} & A word translated by a too general term (undertranslation). Special care was given in order to assure the word used in perturbed text is more general, and incorrect, translation of the original word. & manual & major\\

   4 & & untranslated & The Polish Air Force will eventually be equipped with 32 F-35 Lightning II fighters \textbf{\color{red}manufactured} by Lockheed Martin. 
   
The Polish Air Force will eventually be equipped with 32 F-35 Lightning II fighters \textbf{\color{red}produkowane} by Lockheed Martin. & One word is being left untranslated. We manually assure that each time only one word is left untranslated. & manual & major\\

   5 & & completeness & She is \textbf{\color{red}in custody} pending prosecution and trial; but any witness evidence could be negatively impacted because her image has been widely published.

She is \textbf{\color{red}\_\_\_\_\_} pending prosecution and trial; but any witness evidence could be negatively impacted because her image has been widely published. & One prepositional phrase is being removed. Whenever possible, we remove the shortest prepositional phrase in order to assure that the perturbed sentence is not much shorter than the original translation. & automatic (Stanza) with manual check & major\\
   6 & & addition & \textbf{\color{red}\_\_\_\_\_} Plants look their best when they are in a natural environment, so resist the temptation to remove "just one."

 \textbf{\color{red}Power} plants look their best when they are in a natural environment, so resist the temptation to remove "just one." & One word is being added. We make sure that the added word does not disturb the grammaticality of the sentence but changes the meaning in a significant way. & manual & critical\\
   7 & & antonym & He has been unable to relieve the \textbf{\color{red}pain} with medication, which the competition prohibits competitors from taking.

He has been unable to relieve the \textbf{\color{red}pleasure} with medication, which the competition prohibits competitors from taking. & One word (noun, verb, adj., or adv.) is being changed to its antonym. & manual & critical\\
   8 & & mistranslation - 
   negation & Last month, a presidential committee \textbf{\color{red}recommended} the resignation of the former CEP as part of measures to push the country toward new elections.

Last month, a presidential committee \textbf{\color{red}didn't recommend} the resignation of the former CEP as part of measures to push the country toward new elections.  & Affirmative sentences are being changed into negations. Rare negations are being changed to affirmative sentences. & manual & critical\\
   9 & & mistranslation  - 
   named entity & Late night presenter \textbf{\color{red}Stephen Colbert} welcomed 17-year-old Thunberg to his show on Tuesday and conducted a lengthy interview with the Swede.

Late night presenter \textbf{\color{red}John Oliver} welcomed 17-year-old Thunberg to his show on Tuesday and conducted a lengthy interview with the Swede. & Named entity is replaced with another named entity from the same category (person, geographic location, and organization). & automatic (Stanza) with manual check & critical\\
   10 & & mistranslation - 
   numbers & The Chinese Consulate General in Houston was established in \textbf{\color{red}1979} and is the first Chinese consulate in the United States.

The Chinese Consulate General in Houston was established in \textbf{\color{red}1997} and is the first Chinese consulate in the United States. & A number is being replaced with an incorrect one. Special attention was given to keep the numerals with resonable/common range for the given category (e.g., 0-100 for percentages; 1-12 for months). We also assure that the replacement will not creat illogical sentence (e.g., replacing "1920" with "1940" in "from 1920 to 1930") & manual & critical\\
   11 & & mistranslation - 
   gender & \textbf{\color{red}He} has been unable to relieve the pain with medication, which the competition prohibits competitors from taking.

\textbf{\color{red}She} has been unable to relieve the pain with medication, which the competition prohibits competitors from taking. & Exactly one feminine pronoun in the sentence (such as ``she'' or ``her'') is being with a masculine pronouns (such as ``he'' or ``him'') or vice-versa. This includes reflexive pronouns (i.e., ``him/herself'') and possessive adjectives (i.e., ``his/her''). & automatic with manual check & critical\\
 \midrule
   12 & \multirow{23}{*}{\rotatebox{90}{\LARGE FLUENCY}} & cohesion  & Scientists want to understand how planets have formed \textbf{\color{red}since} a comet collided with Earth long ago, and especially how Earth has formed.

Scientists want to understand how planets have formed \textbf{\color{red}\_\_\_\_\_} a comet collided with Earth long ago, and especially how Earth has formed. & A conjunction, such as ``thus'' or ``therefore'' is removed. Special attention was given to keep the rest of the sentence unperturbed. & automatic (spaCy) with manual check & minor\\
   13 & & grammar - 
   
   pos shift & The U.S. Supreme Court last year blocked the Trump \textbf{\color{red}administration} from including the citizenship question on the 2020 census form.

The U.S. Supreme Court last year blocked the Trump \textbf{\color{red}administrate} from including the citizenship question on the 2020 census form. & Suffix of the word is being changed keeping the root constant (e.g., ``bad'' to ``badly'') which results in the part-of-speech shift. The degree to which the original meaning is affected varies, however, the intended meaning is easily retrivable from the perturbed word. & manual & minor\\
   14 & & grammar - 
   
   order swap & I don't know if you realize that most of the goods imported \textbf{\color{red}into this} country from Central America are duty free.

I don't know if you realize that most of the goods imported \textbf{\color{red}this into} country from Central America are duty free. & Two neighboring words are being swapped to mimic word order error. & automatic (spaCy) & minor\\
   15 & & grammar - 
   
   case & \textbf{\color{red}She} announced that after a break of several years, a Rakoczy horse show will take place again in 2021.

\textbf{\color{red}Her} announced that after a break of several years, a Rakoczy horse show will take place again in 2021. & One pronoun in the sentence is being changed into a different, incorrect, case (e.g., ``he'' to ``him''). & automatic (spaCy) with manual check & minor\\
   16 & & grammar - 
   
   function word & Last month, \textbf{\color{red}a} presidential committee recommended the resignation of the former CEP as part of measures to push the country toward new elections.

Last month, \textbf{\color{red}an} presidential committee recommended the resignation of the former CEP as part of measures to push the country toward new elections. & A preposition or article is being changed into an incorrect one to mimic mistake in function words usage. While most perturbations result in minor mistakes (i.e., the original meaning is easily retrivable) some may be more severe. & automatic with manual check & minor-major\\
   17 & & grammar - 
   
   tense & Cyanuric acid and melamine \textbf{\color{red}were} both found in urine samples of pets who died after eating contaminated pet food.

Cyanuric acid and melamine \textbf{\color{red}are} both found in urine samples of pets who died after eating contaminated pet food. & A tense is being change into an incorrect one. We consider past, present, as well as the future tense (although this may be classified as modal verb in English) & manual & major\\
   18 & & grammar - 
   
   aspect & He \textbf{\color{red}has been} unable to relieve the pain with medication, which the competition prohibits competitors from taking.

He \textbf{\color{red}is being} unable to relieve the pain with medication, which the competition prohibits competitors from taking. & Aspect is being changed to an incorrect one (e.g., perfective to progressive) \textit{without} changing the tense. & manual & major\\
   19 & & grammar - 
   
   interrogative & This is the tenth time since the start of the pandemic that Florida's daily death toll has surpassed 100.

Is this the tenth time since the start of the pandemic that Florida's daily death toll has surpassed 100? & Affirmative mood is being changed to interrogative mood. & manual & major\\
 \midrule
   20 & \multirow{15}{*}{\rotatebox{90}{\Large MIXED}} & omission - 
   
   adj/adv & Rangers \textbf{\color{red}closely} monitor shooters participating in supplemental pest control trials as the trials are monitored and their effectiveness assessed.

Rangers \textbf{\color{red}\_\_\_\_\_} monitor shooters participating in supplemental pest control trials as the trials are monitored and their effectiveness assessed. & An adjective or adverb is being removed. While in most cases this leads to & automatic with manual check & minor-major\\
   21 & & omission  - 
   
   content verb & Catri \textbf{\color{red}said} that 85\% of new coronavirus cases in Belgium last week were under the age of 60.

Catri \textbf{\color{red}\_\_\_\_\_} that 85\% of new coronavirus cases in Belgium last week were under the age of 60. & Content verb is being removed (this excludes auxilary verbs and copulae). & Automatic with manual check & critical\\
   22 & & omission - 
   
   noun & In 1940 he stood up to other government \textbf{\color{red}aristocrats} who wanted to discuss an "agreement" with the Nazis and he very ably won.

In 1940 he stood up to other government \textbf{\color{red}\_\_\_\_\_} who wanted to discuss an "agreement" with the Nazis and he very ably won.

 & Noun, which is not a named entity or a subject, is being removed. We remove the head of the noun phrase including compound nouns. & automatic (spaCy) with manual check & critical\\
   23 & & omission  - 

   subject & His \textbf{\color{red}research} shows that the administration of hormones can accelerate the maturation of the baby's fetal lungs.

His \textbf{\color{red}\_\_\_\_\_} shows that the administration of hormones can accelerate the maturation of the baby's fetal lungs.

 & Subject is being removed. We remove the head of the noun phrase including compound nouns. & automatic (spaCy) with manual check & critical\\
   24 & & omission - 
   named entry & I don't know if you realize that most of the goods imported into this country from \textbf{\color{red}Central America} are duty free.

I don't know if you realize that most of the goods imported into this country from \textbf{\color{red}\_\_\_\_\_} are duty free.
 & Named entity, which is not a subject, is being removed. & automatic (Stanza) with manual check & critical\\
 
\bottomrule
\end{tabular}}
\end{center}
\caption{A full list of perturbations included in \name\ .}
\label{tab:demetr_table_details1}
\end{table*}

\begin{table*}
\small
\begin{center}
\renewcommand*{\arraystretch}{1.5}
\resizebox{\textwidth}{!}{\begin{tabular}{ccp{2cm}p{10cm}p{7cm}p{2cm}c} 
 \toprule
 {\large ID} & {\large Category} & {\large Type} & {\large Example} & {\large Description} & {\large Application} & {\large Error Severity} \\
 \midrule
  25 & \multirow{20}{*}{\rotatebox{90}{\LARGE TYPOGRAPHY}} & spelling - 
   
  misspelled & Scientists want to understand how planets have formed since a comet collided with Earth long ago, and \textbf{\color{red}especially} how Earth has formed.

Scientists want to understand how planets have formed since a comet collided with Earth long ago, and \textbf{\color{red}expecially} how Earth has formed. & One word is being misspelled based on the list of most common misspelled words.\footnote{\url{https://en.wikipedia.org/wiki/Wikipedia:Lists_of_common_misspellings/For_machines}} A word is considered a candidate for misspelling only up to 10 times. & automatic & minor\\
  26 & & spelling - 
   
  char removed & I don't know if you \textbf{\color{red}realize} that most of the goods imported into this country from Central America are duty free.

I don't know if you \textbf{\color{red}realie} that most of the goods imported into this country from Central America are duty free. & A character in a word is being deleted. We consider only nouns, adverbs, adjectives, and verbs as candidates. & automatic & minor\\
  27 & & punctuation - 
   
  removed & When a satellite in space receives a call, it reflects it back almost immediately\textbf{\color{red}.}

When a satellite in space receives a call, it reflects it back almost immediately\textbf{\color{red}\_} & Final punctuation is being removed. & Automatic & minor\\
  28 & & punctuation  - 
   
  added & Comets may have been the source of Earth's water and organic matter that can form proteins and sustain life.

Comets may have been the source of Earth's\textbf{\color{red},} water and organic matter that can form proteins and sustain life. & A punctuation is being added. & Automatic & minor\\
  29 & & tokenized & At 9:30 a.m. on July 26, the reporter saw at the scene of Jiangkouhe Lianxu that the local area had made various preparations before flood distribution.

At 9:30 a.m. on July 2\textbf{\color{red}6 , t}he reporter saw at the scene of Jiangkouhe Lianxu that the local area had made various preparations before flood distribution . & The sentence is tokenized. & Automatic & minor\\
  30 & & lowercases - 
   
  whole & For example, U.S. citizens in the Middle East may face different situations than Europeans or Arabs.
   
for example, u.s. citizens in the middle east may face different situations than europeans or arabs. & The entire sentence is lowercased. & Automatic & minor\\
  31 & & lowercases - 
   
  first word & For example, U.S. citizens in the Middle East may face different situations than Europeans or Arabs.

for example, U.S. citizens in the Middle East may face different situations than Europeans or Arabs. & The first word in a sentence is lowercased. & Automatic & minor\\
 \midrule
  32 & \multirow{10}{*}{\rotatebox{90}{\LARGE BASELINE}} & empty  & In the next two instances they have proved Freudenberg the right, but the opposite part continues to fight today. & Empty string (since most automatic metrics will not allow an empty string we pass a full stop instead). & Automatic & base\\
  33 & & different & I don't know if you realize that most of the goods imported into this country from Central America are duty free.

\textbf{\color{red}It was the last game for the All Blacks, who had won the trophy two weeks earlier.} & Unrelated translation. & Automatic & base\\
  34 & & unintelligible & Cyanuric acid and melamine were both found in urine samples of pets who died after eating contaminated pet food.

Pets urine in of and acid were both died melamine found pet after who eating food contaminated cyanuric samples. & Shuffled words. & Automatic & base\\
  35 & & reference & Last month, a presidential committee recommended the resignation of the former CEP as part of measures to push the country toward new elections.

Last month a presidential commission recommended the prior CEP's resignation as part of a package of measures to move the country towards new elections. & Reference passed as the translation. & Automatic & base\\
\bottomrule
\end{tabular}}
\end{center}
\caption{Table~\ref{tab:demetr_table_details1} continued.}
\label{tab:demetr_table_details2}
\end{table*}

\begin{table*}
\small
\begin{center}
\resizebox{\columnwidth}{!}{\begin{tabular}{cllcrrrr} 
 \toprule
\multicolumn{1}{l}{} & & & \multicolumn{1}{l}{} & \multicolumn{3}{l}{Welsch \textit{t}-test} \\
\multicolumn{1}{l}{\multirow{-2}{*}{ID}}  & \multirow{-2}{*}{perturbation} & \multirow{-2}{*}{metric} & \multicolumn{1}{l}{\multirow{-2}{*}{type}} & \multicolumn{1}{l}{\textit{t}} & \multicolumn{1}{l}{\textit{p}-val} & \multicolumn{1}{l}{df}  & \multirow{-2}{*}{accuracy} \\
 \midrule

 &  & \textsc{Bleu} & string & 2.98 & 0.003 & 1,992.12 & 93.2\% \\
 &  & \textsc{Meteor} & string & 0.44 & 0.662 & 1,997.34 & 85.4\% \\
 &  & \textsc{ChrF} & string & 1.00 & 0.316 & 1,997.04 & 89.9\% \\
 &  & \textsc{ChrF2} & string & 0.96 & 0.337 & 1,997.23 & 92.7\% \\
 &  & \textsc{TER} & string & -3.88 & \textless{}0.001 & 1,996.94 & 77.7\% \\
 &  & \textsc{CER} & string & -5.61 & \textless{}0.001 & 1,995.16 & 88.8\% \\
 &  & \textsc{Rouge2} & string & 1.69 & 0.092 & 1,996.37 & 99.7\% \\
 &  & \textsc{BERTScore} & learned & 8.43 & \textless{}0.001 & 1,997.55 & 97.5\% \\
 &  & \textsc{Comet-QE} & learned & 21.33 & \textless{}0.001 & 1,991.48 & 99.1\% \\
 &  & \textsc{Comet} & learned & 22.46 & \textless{}0.001 & 1,973.93 & 99.1\% \\
 &  & \textsc{Bleurt20} & learned & 24.43 & \textless{}0.001 & 1,996.98 & 99.0\% \\
 &  & \textsc{Prism-QE} & learned & 6.86 & \textless{}0.001 & 1,989.71 & 98.9\% \\
 &  & \textsc{Prism} & learned & 9.61 & \textless{}0.001 & 1,996.17 & 99.9\% \\
\multirow{-14}{*}{\textbf{1}} &  \multirow{-14}{*}{addition (repetition)} & BARTScore & learned & 1.49 & 0.137 & 1,997.92 & 78.0\% \\
\midrule
 &  & \textsc{Bleu} & string & 6.47 & \textless{}0.001 & 1,960.90 & 95.7\% \\
 &  & \textsc{Meteor} & string & 1.63 & 0.104 & 1,997.00 & 85.8\% \\
 &  & \textsc{ChrF} & string & 3.88 & \textless{}0.001 & 1,992.23 & 97.6\% \\
 &  & \textsc{ChrF2} & string & 3.53 & \textless{}0.001 & 1,993.49 & 98.5\% \\
 &  & \textsc{TER} & string & -13.93 & \textless{}0.001 & 1,996.08 & 95.2\% \\
 &  & \textsc{CER} & string & -18.65 & \textless{}0.001 & 1,992.80 & 96.1\% \\
 &  & \textsc{Rouge2} & string & 4.92 & \textless{}0.001 & 1,984.62 & 99.7\% \\
 &  & \textsc{BERTScore} & learned & 22.91 & \textless{}0.001 & 1,990.44 & 99.9\% \\
 &  & \textsc{Comet-QE} & learned & 34.03 & \textless{}0.001 & 1,982.14 & 100.0\% \\
 &  & \textsc{Comet} & learned & 42.55 & \textless{}0.001 & 1,955.19 & 100.0\% \\
 &  & \textsc{Bleurt20} & learned & 50.88 & \textless{}0.001 & 1,991.41 & 99.9\% \\
 &  & \textsc{Prism-QE} & learned & 19.02 & \textless{}0.001 & 1,945.31 & 98.7\% \\
 &  & \textsc{Prism} & learned & 27.18 & \textless{}0.001 & 1,994.44 & 100.0\% \\
\multirow{-14}{*}{\textbf{2}} &  \multirow{-14}{*}{addition (repetition)} & BARTScore & learned & 5.76 & \textless{}0.001 & 1,997.69 & 95.0\% \\
\midrule
 &  & \textsc{Bleu} & string & 5.11 & \textless{}0.001 & 1,767.45 & 69.5\% \\
 &  & \textsc{Meteor} & string & 5.12 & \textless{}0.001 & 1,785.42 & 66.3\% \\
 &  & \textsc{ChrF} & string & 8.67 & \textless{}0.001 & 1,777.75 & 89.7\% \\
 &  & \textsc{ChrF2} & string & 7.93 & \textless{}0.001 & 1,777.56 & 89.3\% \\
 &  & \textsc{TER} & string & -3.30 & \textless{}0.001 & 1,784.05 & 53.2\% \\
 &  & \textsc{CER} & string & -4.05 & \textless{}0.001 & 1,780.32 & 77.9\% \\
 &  & \textsc{Rouge2} & string & 5.73 & \textless{}0.001 & 1,776.50 & 63.3\% \\
 &  & \textsc{BERTScore} & learned & 8.86 & \textless{}0.001 & 1,784.44 & 93.6\% \\
 &  & \textsc{Comet-QE} & learned & 4.24 & \textless{}0.001 & 1,785.74 & 78.1\% \\
 &  & \textsc{Comet} & learned & 7.10 & \textless{}0.001 & 1,784.74 & 91.2\% \\
 &  & \textsc{Bleurt20} & learned & 13.80 & \textless{}0.001 & 1,786.00 & 92.7\% \\
 &  & \textsc{Prism-QE} & learned & 4.46 & \textless{}0.001 & 1,781.82 & 94.4\% \\
 &  & \textsc{Prism} & learned & 10.40 & \textless{}0.001 & 1,785.87 & 95.7\% \\
\multirow{-14}{*}{\textbf{3}} &  \multirow{-14}{*}{hypernym (undertranslation)} & BARTScore & learned & 6.42 & \textless{}0.001 & 1,781.16 & 90.5\% \\
\midrule
 &  & \textsc{Bleu} & string & 5.70 & \textless{}0.001 & 1,973.31 & 73.1\% \\
 &  & \textsc{Meteor} & string & 6.18 & \textless{}0.001 & 1,995.59 & 72.5\% \\
 &  & \textsc{ChrF} & string & 9.22 & \textless{}0.001 & 1,989.48 & 95.2\% \\
 &  & \textsc{ChrF2} & string & 8.56 & \textless{}0.001 & 1,988.24 & 95.3\% \\
 &  & \textsc{TER} & string & -3.82 & \textless{}0.001 & 1,994.43 & 58.3\% \\
 &  & \textsc{CER} & string & -4.53 & \textless{}0.001 & 1,992.88 & 83.5\% \\
 &  & \textsc{Rouge2} & string & 6.35 & \textless{}0.001 & 1,984.97 & 68.4\% \\
 &  & \textsc{BERTScore} & learned & 36.69 & \textless{}0.001 & 1,824.17 & 99.8\% \\
 &  & \textsc{Comet-QE} & learned & 27.31 & \textless{}0.001 & 1,994.98 & 98.3\% \\
 &  & \textsc{Comet} & learned & 26.78 & \textless{}0.001 & 1,997.69 & 99.2\% \\
 &  & \textsc{Bleurt20} & learned & 24.84 & \textless{}0.001 & 1,822.75 & 99.1\% \\
 &  & \textsc{Prism-QE} & learned & 10.38 & \textless{}0.001 & 1,993.90 & 97.6\% \\
 &  & \textsc{Prism} & learned & 16.62 & \textless{}0.001 & 1,988.33 & 99.8\% \\
\multirow{-14}{*}{\textbf{4}} &  \multirow{-14}{*}{untranslated} & BARTScore & learned & 8.91 & \textless{}0.001 & 1,991.19 & 90.8\% \\
\midrule
 &  & \textsc{Bleu} & string & 9.94 & \textless{}0.001 & 1,748.59 & 79.2\% \\
 &  & \textsc{Meteor} & string & 17.83 & \textless{}0.001 & 1,777.96 & 89.4\% \\
 &  & \textsc{ChrF} & string & 15.88 & \textless{}0.001 & 1,777.31 & 93.1\% \\
 &  & \textsc{ChrF2} & string & 15.77 & \textless{}0.001 & 1,778.00 & 94.0\% \\
 &  & \textsc{TER} & string & -9.79 & \textless{}0.001 & 1,715.36 & 76.1\% \\
 &  & \textsc{CER} & string & -8.48 & \textless{}0.001 & 1,715.24 & 73.6\% \\
 &  & \textsc{Rouge2} & string & 8.42 & \textless{}0.001 & 1,775.53 & 78.7\% \\
 &  & \textsc{BERTScore} & learned & 18.07 & \textless{}0.001 & 1,770.58 & 94.2\% \\
 &  & \textsc{Comet-QE} & learned & 6.98 & \textless{}0.001 & 1,777.40 & 80.2\% \\
 &  & \textsc{Comet} & learned & 13.84 & \textless{}0.001 & 1,777.28 & 95.5\% \\
 &  & \textsc{Bleurt20} & learned & 25.70 & \textless{}0.001 & 1,579.21 & 97.1\% \\
 &  & \textsc{Prism-QE} & learned & 6.14 & \textless{}0.001 & 1,776.54 & 92.6\% \\
 &  & \textsc{Prism} & learned & 19.76 & \textless{}0.001 & 1,743.67 & 96.1\% \\
\multirow{-14}{*}{\textbf{5}} &  \multirow{-14}{*}{completeness (omitted pp)} & BARTScore & learned & 19.56 & \textless{}0.001 & 1,670.43 & 96.2\% \\
\midrule
 &  & \textsc{Bleu} & string & 4.84 & \textless{}0.001 & 1,979.16 & 93.0\% \\
 &  & \textsc{Meteor} & string & 1.53 & 0.127 & 1,997.87 & 99.0\% \\
 &  & \textsc{ChrF} & string & 3.12 & 0.002 & 1,992.92 & 89.4\% \\
 &  & \textsc{ChrF2} & string & 3.06 & 0.002 & 1,993.62 & 91.8\% \\
 &  & \textsc{TER} & string & -4.74 & \textless{}0.001 & 1,997.95 & 80.3\% \\
 &  & \textsc{CER} & string & -6.24 & \textless{}0.001 & 1,996.99 & 92.9\% \\
 &  & \textsc{Rouge2} & string & 4.90 & \textless{}0.001 & 1,987.76 & 99.8\% \\
 &  & \textsc{BERTScore} & learned & 9.54 & \textless{}0.001 & 1,994.95 & 98.5\% \\
 &  & \textsc{Comet-QE} & learned & 4.78 & \textless{}0.001 & 1,997.76 & 69.7\% \\
 &  & \textsc{Comet} & learned & 9.25 & \textless{}0.001 & 1,996.06 & 93.3\% \\
 &  & \textsc{Bleurt20} & learned & 19.09 & \textless{}0.001 & 1,996.53 & 97.1\% \\
 &  & \textsc{Prism-QE} & learned & 7.13 & \textless{}0.001 & 1,991.72 & 98.3\% \\
 &  & \textsc{Prism} & learned & 13.58 & \textless{}0.001 & 1,995.57 & 99.9\% \\
\multirow{-14}{*}{\textbf{6}} &  \multirow{-14}{*}{addition} & BARTScore & learned & 5.56 & \textless{}0.001 & 1,997.62 & 94.1\% \\

\bottomrule
\end{tabular}}
\end{center}
\label{tab:significance1}
\end{table*}

\begin{table*}
\small
\begin{center}
\resizebox{\columnwidth}{!}{\begin{tabular}{cllcrrrr} 
 \toprule
\multicolumn{1}{l}{} & & & \multicolumn{1}{l}{} & \multicolumn{3}{l}{Welsch \textit{t}-test} \\
\multicolumn{1}{l}{\multirow{-2}{*}{ID}}  & \multirow{-2}{*}{perturbation} & \multirow{-2}{*}{metric} & \multicolumn{1}{l}{\multirow{-2}{*}{type}} & \multicolumn{1}{l}{\textit{t}} & \multicolumn{1}{l}{\textit{p}-val} & \multicolumn{1}{l}{df}  & \multirow{-2}{*}{accuracy} \\

\midrule
 &  & \textsc{Bleu} & string & 5.47 & \textless{}0.001 & 1,949.90 & 65.4\% \\
 &  & \textsc{Meteor} & string & 5.67 & \textless{}0.001 & 1,963.08 & 68.9\% \\
 &  & \textsc{ChrF} & string & 7.11 & \textless{}0.001 & 1,953.68 & 83.8\% \\
 &  & \textsc{ChrF2} & string & 6.81 & \textless{}0.001 & 1,954.25 & 84.0\% \\
 &  & \textsc{TER} & string & -3.49 & \textless{}0.001 & 1,961.27 & 64.9\% \\
 &  & \textsc{CER} & string & -3.34 & \textless{}0.001 & 1,959.91 & 76.8\% \\
 &  & \textsc{Rouge2} & string & 6.10 & \textless{}0.001 & 1,952.61 & 57.7\% \\
 &  & \textsc{BERTScore} & learned & 11.52 & \textless{}0.001 & 1,962.23 & 98.3\% \\
 &  & \textsc{Comet-QE} & learned & 6.61 & \textless{}0.001 & 1,963.91 & 83.7\% \\
 &  & \textsc{Comet} & learned & 12.11 & \textless{}0.001 & 1,955.91 & 97.0\% \\
 &  & \textsc{Bleurt20} & learned & 24.45 & \textless{}0.001 & 1,939.39 & 98.7\% \\
 &  & \textsc{Prism-QE} & learned & 6.39 & \textless{}0.001 & 1,954.01 & 96.7\% \\
 &  & \textsc{Prism} & learned & 13.85 & \textless{}0.001 & 1,960.97 & 99.2\% \\
\multirow{-14}{*}{\textbf{7}} &  \multirow{-14}{*}{antonym} & BARTScore & learned & 7.34 & \textless{}0.001 & 1,963.98 & 96.9\% \\
\midrule
 &  & \textsc{Bleu} & string & 7.32 & \textless{}0.001 & 1,961.08 & 91.3\% \\
 &  & \textsc{Meteor} & string & 3.36 & \textless{}0.001 & 1,995.95 & 94.7\% \\
 &  & \textsc{ChrF} & string & 4.88 & \textless{}0.001 & 1,990.62 & 90.8\% \\
 &  & \textsc{ChrF2} & string & 5.14 & \textless{}0.001 & 1,990.35 & 94.6\% \\
 &  & \textsc{TER} & string & -8.26 & \textless{}0.001 & 1,995.94 & 89.6\% \\
 &  & \textsc{CER} & string & -5.29 & \textless{}0.001 & 1,995.38 & 91.0\% \\
 &  & \textsc{Rouge2} & string & 7.67 & \textless{}0.001 & 1,979.28 & 96.3\% \\
 &  & \textsc{BERTScore} & learned & 15.60 & \textless{}0.001 & 1,995.96 & 99.6\% \\
 &  & \textsc{Comet-QE} & learned & 6.86 & \textless{}0.001 & 1,995.73 & 83.7\% \\
 &  & \textsc{Comet} & learned & 18.35 & \textless{}0.001 & 1,991.67 & 99.4\% \\
 &  & \textsc{Bleurt20} & learned & 41.51 & \textless{}0.001 & 1,987.24 & 99.8\% \\
 &  & \textsc{Prism-QE} & learned & 8.43 & \textless{}0.001 & 1,977.03 & 96.1\% \\
 &  & \textsc{Prism} & learned & 16.17 & \textless{}0.001 & 1,994.10 & 99.8\% \\
\multirow{-14}{*}{\textbf{8}} &  \multirow{-14}{*}{mistranslation - negation} & BARTScore & learned & 9.44 & \textless{}0.001 & 1,988.26 & 98.5\% \\
\midrule
 &  & \textsc{Bleu} & string & 7.00 & \textless{}0.001 & 1,339.02 & 90.5\% \\
 &  & \textsc{Meteor} & string & 8.29 & \textless{}0.001 & 1,364.57 & 89.3\% \\
 &  & \textsc{ChrF} & string & 12.47 & \textless{}0.001 & 1,362.50 & 98.7\% \\
 &  & \textsc{ChrF2} & string & 11.54 & \textless{}0.001 & 1,361.48 & 98.7\% \\
 &  & \textsc{TER} & string & -7.14 & \textless{}0.001 & 1,361.50 & 83.8\% \\
 &  & \textsc{CER} & string & -7.58 & \textless{}0.001 & 1,358.56 & 89.9\% \\
 &  & \textsc{Rouge2} & string & 8.73 & \textless{}0.001 & 1,350.27 & 87.7\% \\
 &  & \textsc{BERTScore} & learned & 25.35 & \textless{}0.001 & 1,358.26 & 99.1\% \\
 &  & \textsc{Comet-QE} & learned & 7.14 & \textless{}0.001 & 1,365.67 & 85.4\% \\
 &  & \textsc{Comet} & learned & 18.20 & \textless{}0.001 & 1,363.20 & 98.8\% \\
 &  & \textsc{Bleurt20} & learned & 43.02 & \textless{}0.001 & 1,279.18 & 100.0\% \\
 &  & \textsc{Prism-QE} & learned & 12.00 & \textless{}0.001 & 1,331.69 & 95.3\% \\
 &  & \textsc{Prism} & learned & 30.02 & \textless{}0.001 & 1,348.23 & 99.7\% \\
\multirow{-14}{*}{\textbf{9}} &  \multirow{-14}{*}{mistranslation - named entry} & BARTScore & learned & 24.24 & \textless{}0.001 & 1,336.08 & 100.0\% \\
\midrule
 &  & \textsc{Bleu} & string & 4.44 & \textless{}0.001 & 734.86 & 89.0\% \\
 &  & \textsc{Meteor} & string & 3.97 & \textless{}0.001 & 741.65 & 79.6\% \\
 &  & \textsc{ChrF} & string & 2.99 & 0.003 & 740.92 & 96.5\% \\
 &  & \textsc{ChrF2} & string & 3.47 & \textless{}0.001 & 740.45 & 96.5\% \\
 &  & \textsc{TER} & string & -2.61 & 0.009 & 741.63 & 79.8\% \\
 &  & \textsc{CER} & string & -0.82 & 0.415 & 741.83 & 80.4\% \\
 &  & \textsc{Rouge2} & string & 4.90 & \textless{}0.001 & 738.46 & 82.5\% \\
 &  & \textsc{BERTScore} & learned & 2.05 & 0.041 & 742.00 & 98.9\% \\
 &  & \textsc{Comet-QE} & learned & 0.16 & 0.871 & 741.99 & 53.2\% \\
 &  & \textsc{Comet} & learned & 0.73 & 0.463 & 741.82 & 80.4\% \\
 &  & \textsc{Bleurt20} & learned & 9.18 & \textless{}0.001 & 739.98 & 98.7\% \\
 &  & \textsc{Prism-QE} & learned & 4.38 & \textless{}0.001 & 741.10 & 99.5\% \\
 &  & \textsc{Prism} & learned & 8.46 & \textless{}0.001 & 741.89 & 100.0\% \\
\multirow{-14}{*}{\textbf{10}} &  \multirow{-14}{*}{mistranslation - numbers} & BARTScore & learned & 7.21 & \textless{}0.001 & 741.99 & 99.5\% \\
\midrule
 &  & \textsc{Bleu} & string & 2.17 & 0.031 & 221.29 & 84.1\% \\
 &  & \textsc{Meteor} & string & 2.13 & 0.035 & 223.97 & 83.2\% \\
 &  & \textsc{ChrF} & string & 1.08 & 0.283 & 223.80 & 87.6\% \\
 &  & \textsc{ChrF2} & string & 1.38 & 0.169 & 223.72 & 90.3\% \\
 &  & \textsc{TER} & string & -1.56 & 0.120 & 223.87 & 83.2\% \\
 &  & \textsc{CER} & string & -0.49 & 0.628 & 223.98 & 85.0\% \\
 &  & \textsc{Rouge2} & string & 2.27 & 0.024 & 221.81 & 72.6\% \\
 &  & \textsc{BERTScore} & learned & 1.95 & 0.052 & 223.87 & 99.1\% \\
 &  & \textsc{Comet-QE} & learned & 1.35 & 0.178 & 223.51 & 61.9\% \\
 &  & \textsc{Comet} & learned & 4.05 & \textless{}0.001 & 222.69 & 96.5\% \\
 &  & \textsc{Bleurt20} & learned & 8.41 & \textless{}0.001 & 223.05 & 99.1\% \\
 &  & \textsc{Prism-QE} & learned & 1.09 & 0.277 & 223.93 & 97.3\% \\
 &  & \textsc{Prism} & learned & 3.16 & 0.002 & 223.97 & 100.0\% \\
\multirow{-14}{*}{\textbf{11}} &  \multirow{-14}{*}{mistranslation - gender} & BARTScore & learned & 1.45 & 0.148 & 223.97 & 96.5\% \\
\midrule
 &  & \textsc{Bleu} & string & 4.43 & \textless{}0.001 & 1,441.98 & 75.8\% \\
 &  & \textsc{Meteor} & string & 5.13 & \textless{}0.001 & 1,449.88 & 78.1\% \\
 &  & \textsc{ChrF} & string & 4.50 & \textless{}0.001 & 1,448.17 & 85.3\% \\
 &  & \textsc{ChrF2} & string & 4.69 & \textless{}0.001 & 1,448.04 & 84.6\% \\
 &  & \textsc{TER} & string & -2.33 & 0.020 & 1,444.72 & 57.9\% \\
 &  & \textsc{CER} & string & -1.88 & 0.060 & 1,443.24 & 70.4\% \\
 &  & \textsc{Rouge2} & string & 4.26 & \textless{}0.001 & 1,447.85 & 62.0\% \\
 &  & \textsc{BERTScore} & learned & 9.89 & \textless{}0.001 & 1,448.55 & 93.9\% \\
 &  & \textsc{Comet-QE} & learned & 7.03 & \textless{}0.001 & 1,448.46 & 89.7\% \\
 &  & \textsc{Comet} & learned & 8.33 & \textless{}0.001 & 1,448.86 & 93.8\% \\
 &  & \textsc{Bleurt20} & learned & 12.68 & \textless{}0.001 & 1,448.58 & 95.0\% \\
 &  & \textsc{Prism-QE} & learned & 5.04 & \textless{}0.001 & 1,449.60 & 97.8\% \\
 &  & \textsc{Prism} & learned & 8.57 & \textless{}0.001 & 1,449.92 & 96.6\% \\
\multirow{-14}{*}{\textbf{12}} &  \multirow{-14}{*}{cohesion} & BARTScore & learned & 3.25 & 0.001 & 1,449.46 & 83.1\% \\

\bottomrule
\end{tabular}}
\end{center}
\label{tab:significance2}
\end{table*}

\begin{table*}
\small
\begin{center}
\resizebox{\columnwidth}{!}{\begin{tabular}{cllcrrrr} 
 \toprule
\multicolumn{1}{l}{} & & & \multicolumn{1}{l}{} & \multicolumn{3}{l}{Welsch \textit{t}-test} \\
\multicolumn{1}{l}{\multirow{-2}{*}{ID}}  & \multirow{-2}{*}{perturbation} & \multirow{-2}{*}{metric} & \multicolumn{1}{l}{\multirow{-2}{*}{type}} & \multicolumn{1}{l}{\textit{t}} & \multicolumn{1}{l}{\textit{p}-val} & \multicolumn{1}{l}{df}  & \multirow{-2}{*}{accuracy} \\

\midrule
 &  & \textsc{Bleu} & string & 5.03 & \textless{}0.001 & 1,972.02 & 63.3\% \\
 &  & \textsc{Meteor} & string & 2.01 & 0.045 & 1,989.94 & 29.3\% \\
 &  & \textsc{ChrF} & string & 3.82 & \textless{}0.001 & 1,984.67 & 85.5\% \\
 &  & \textsc{ChrF2} & string & 4.29 & \textless{}0.001 & 1,983.23 & 85.9\% \\
 &  & \textsc{TER} & string & -3.12 & 0.002 & 1,987.02 & 56.0\% \\
 &  & \textsc{CER} & string & -1.84 & 0.067 & 1,989.04 & 79.4\% \\
 &  & \textsc{Rouge2} & string & 5.75 & \textless{}0.001 & 1,974.55 & 60.3\% \\
 &  & \textsc{BERTScore} & learned & 14.08 & \textless{}0.001 & 1,983.09 & 96.3\% \\
 &  & \textsc{Comet-QE} & learned & 12.20 & \textless{}0.001 & 1,987.47 & 95.0\% \\
 &  & \textsc{Comet} & learned & 11.36 & \textless{}0.001 & 1,988.92 & 97.0\% \\
 &  & \textsc{Bleurt20} & learned & 13.27 & \textless{}0.001 & 1,987.72 & 96.6\% \\
 &  & \textsc{Prism-QE} & learned & 7.02 & \textless{}0.001 & 1,989.26 & 98.8\% \\
 &  & \textsc{Prism} & learned & 9.68 & \textless{}0.001 & 1,990.00 & 98.8\% \\
\multirow{-14}{*}{\textbf{13}} &  \multirow{-14}{*}{grammar - pos shift} & BARTScore & learned & 1.60 & 0.110 & 1,989.99 & 72.9\% \\
\midrule
 &  & \textsc{Bleu} & string & 7.51 & \textless{}0.001 & 1,968.29 & 72.9\% \\
 &  & \textsc{Meteor} & string & 2.99 & 0.003 & 1,997.13 & 69.9\% \\
 &  & \textsc{ChrF} & string & 6.05 & \textless{}0.001 & 1,984.74 & 82.2\% \\
 &  & \textsc{ChrF2} & string & 5.92 & \textless{}0.001 & 1,984.84 & 82.3\% \\
 &  & \textsc{TER} & string & -3.65 & \textless{}0.001 & 1,993.55 & 62.9\% \\
 &  & \textsc{CER} & string & -4.35 & \textless{}0.001 & 1,985.78 & 74.2\% \\
 &  & \textsc{Rouge2} & string & 9.80 & \textless{}0.001 & 1,969.34 & 76.8\% \\
 &  & \textsc{BERTScore} & learned & 18.87 & \textless{}0.001 & 1,994.91 & 98.6\% \\
 &  & \textsc{Comet-QE} & learned & 18.47 & \textless{}0.001 & 1,996.51 & 97.8\% \\
 &  & \textsc{Comet} & learned & 16.44 & \textless{}0.001 & 1,996.53 & 98.4\% \\
 &  & \textsc{Bleurt20} & learned & 24.50 & \textless{}0.001 & 1,983.00 & 98.5\% \\
 &  & \textsc{Prism-QE} & learned & 11.91 & \textless{}0.001 & 1,995.77 & 99.3\% \\
 &  & \textsc{Prism} & learned & 16.57 & \textless{}0.001 & 1,997.94 & 99.3\% \\
\multirow{-14}{*}{\textbf{14}} &  \multirow{-14}{*}{grammar - order swap} & BARTScore & learned & 2.74 & 0.006 & 1,997.81 & 78.6\% \\
\midrule
 &  & \textsc{Bleu} & string & 3.05 & 0.002 & 677.34 & 68.5\% \\
 &  & \textsc{Meteor} & string & 2.80 & 0.005 & 683.76 & 63.0\% \\
 &  & \textsc{ChrF} & string & 1.91 & 0.056 & 683.25 & 86.9\% \\
 &  & \textsc{ChrF2} & string & 2.30 & 0.022 & 682.89 & 88.6\% \\
 &  & \textsc{TER} & string & -2.14 & 0.032 & 683.12 & 67.3\% \\
 &  & \textsc{CER} & string & -1.40 & 0.161 & 683.88 & 83.4\% \\
 &  & \textsc{Rouge2} & string & 3.44 & \textless{}0.001 & 679.73 & 63.8\% \\
 &  & \textsc{BERTScore} & learned & 7.35 & \textless{}0.001 & 682.78 & 99.7\% \\
 &  & \textsc{Comet-QE} & learned & 8.25 & \textless{}0.001 & 683.95 & 97.7\% \\
 &  & \textsc{Comet} & learned & 8.12 & \textless{}0.001 & 683.13 & 98.8\% \\
 &  & \textsc{Bleurt20} & learned & 9.00 & \textless{}0.001 & 684.00 & 99.1\% \\
 &  & \textsc{Prism-QE} & learned & 5.32 & \textless{}0.001 & 683.76 & 99.7\% \\
 &  & \textsc{Prism} & learned & 6.31 & \textless{}0.001 & 684.00 & 99.7\% \\
\multirow{-14}{*}{\textbf{15}} &  \multirow{-14}{*}{grammar - case} & BARTScore & learned & 0.75 & 0.453 & 683.98 & 73.8\% \\
\midrule
 &  & \textsc{Bleu} & string & 6.07 & \textless{}0.001 & 1,943.76 & 70.1\% \\
 &  & \textsc{Meteor} & string & 5.22 & \textless{}0.001 & 1,963.26 & 67.3\% \\
 &  & \textsc{ChrF} & string & 3.74 & \textless{}0.001 & 1,959.17 & 83.2\% \\
 &  & \textsc{ChrF2} & string & 4.39 & \textless{}0.001 & 1,958.09 & 85.2\% \\
 &  & \textsc{TER} & string & -3.67 & \textless{}0.001 & 1,963.43 & 76.4\% \\
 &  & \textsc{CER} & string & -2.19 & 0.028 & 1,965.20 & 81.1\% \\
 &  & \textsc{Rouge2} & string & 6.47 & \textless{}0.001 & 1,949.80 & 69.9\% \\
 &  & \textsc{BERTScore} & learned & 11.34 & \textless{}0.001 & 1,961.19 & 97.8\% \\
 &  & \textsc{Comet-QE} & learned & 9.19 & \textless{}0.001 & 1,962.86 & 88.6\% \\
 &  & \textsc{Comet} & learned & 8.70 & \textless{}0.001 & 1,965.91 & 91.8\% \\
 &  & \textsc{Bleurt20} & learned & 13.79 & \textless{}0.001 & 1,960.54 & 93.8\% \\
 &  & \textsc{Prism-QE} & learned & 7.26 & \textless{}0.001 & 1,965.92 & 99.8\% \\
 &  & \textsc{Prism} & learned & 10.49 & \textless{}0.001 & 1,965.89 & 99.3\% \\
\multirow{-14}{*}{\textbf{16}} &  \multirow{-14}{*}{grammar - function word} & BARTScore & learned & 1.53 & 0.126 & 1,965.88 & 78.8\% \\
\midrule
 &  & \textsc{Bleu} & string & 6.19 & \textless{}0.001 & 1,946.80 & 78.7\% \\
 &  & \textsc{Meteor} & string & 3.66 & \textless{}0.001 & 1,973.48 & 66.7\% \\
 &  & \textsc{ChrF} & string & 4.56 & \textless{}0.001 & 1,965.71 & 89.0\% \\
 &  & \textsc{ChrF2} & string & 5.08 & \textless{}0.001 & 1,964.10 & 89.7\% \\
 &  & \textsc{TER} & string & -5.37 & \textless{}0.001 & 1,971.28 & 82.5\% \\
 &  & \textsc{CER} & string & -2.67 & 0.008 & 1,973.18 & 82.8\% \\
 &  & \textsc{Rouge2} & string & 7.06 & \textless{}0.001 & 1,949.72 & 81.2\% \\
 &  & \textsc{BERTScore} & learned & 6.82 & \textless{}0.001 & 1,971.00 & 96.5\% \\
 &  & \textsc{Comet-QE} & learned & 3.32 & \textless{}0.001 & 1,973.76 & 82.8\% \\
 &  & \textsc{Comet} & learned & 4.03 & \textless{}0.001 & 1,970.80 & 92.7\% \\
 &  & \textsc{Bleurt20} & learned & 10.30 & \textless{}0.001 & 1,965.99 & 94.5\% \\
 &  & \textsc{Prism-QE} & learned & 4.00 & \textless{}0.001 & 1,970.72 & 96.2\% \\
 &  & \textsc{Prism} & learned & 7.55 & \textless{}0.001 & 1,972.27 & 98.5\% \\
\multirow{-14}{*}{\textbf{17}} &  \multirow{-14}{*}{grammar - tense} & BARTScore & learned & 2.92 & 0.004 & 1,973.99 & 91.0\% \\
\midrule
 &  & \textsc{Bleu} & string & 6.30 & \textless{}0.001 & 1,945.00 & 92.3\% \\
 &  & \textsc{Meteor} & string & 2.25 & 0.025 & 1,972.00 & 84.6\% \\
 &  & \textsc{ChrF} & string & 5.14 & \textless{}0.001 & 1,959.98 & 88.6\% \\
 &  & \textsc{ChrF2} & string & 5.37 & \textless{}0.001 & 1,960.69 & 90.6\% \\
 &  & \textsc{TER} & string & -6.95 & \textless{}0.001 & 1,968.84 & 84.7\% \\
 &  & \textsc{CER} & string & -5.03 & \textless{}0.001 & 1,969.92 & 91.4\% \\
 &  & \textsc{Rouge2} & string & 7.04 & \textless{}0.001 & 1,952.27 & 96.6\% \\
 &  & \textsc{BERTScore} & learned & 7.86 & \textless{}0.001 & 1,969.14 & 97.1\% \\
 &  & \textsc{Comet-QE} & learned & 3.12 & 0.002 & 1,972.00 & 78.0\% \\
 &  & \textsc{Comet} & learned & 3.60 & \textless{}0.001 & 1,969.97 & 89.4\% \\
 &  & \textsc{Bleurt20} & learned & 9.15 & \textless{}0.001 & 1,953.54 & 94.2\% \\
 &  & \textsc{Prism-QE} & learned & 4.49 & \textless{}0.001 & 1,967.57 & 95.3\% \\
 &  & \textsc{Prism} & learned & 7.57 & \textless{}0.001 & 1,969.38 & 97.3\% \\
\multirow{-14}{*}{\textbf{18}} &  \multirow{-14}{*}{grammar - aspect} & BARTScore & learned & 0.85 & 0.394 & 1,971.98 & 74.8\% \\

\bottomrule
\end{tabular}}
\end{center}
\label{tab:significance3}
\end{table*}

\begin{table*}
\small
\begin{center}
\resizebox{\columnwidth}{!}{\begin{tabular}{cllcrrrr} 
 \toprule
\multicolumn{1}{l}{} & & & \multicolumn{1}{l}{} & \multicolumn{3}{l}{Welsch \textit{t}-test} \\
\multicolumn{1}{l}{\multirow{-2}{*}{ID}}  & \multirow{-2}{*}{perturbation} & \multirow{-2}{*}{metric} & \multicolumn{1}{l}{\multirow{-2}{*}{type}} & \multicolumn{1}{l}{\textit{t}} & \multicolumn{1}{l}{\textit{p}-val} & \multicolumn{1}{l}{df}  & \multirow{-2}{*}{accuracy} \\

\midrule
 &  & \textsc{Bleu} & string & 11.11 & \textless{}0.001 & 1,851.84 & 97.4\% \\
 &  & \textsc{Meteor} & string & 6.80 & \textless{}0.001 & 1,917.78 & 91.4\% \\
 &  & \textsc{ChrF} & string & 6.51 & \textless{}0.001 & 1,911.45 & 94.1\% \\
 &  & \textsc{ChrF2} & string & 8.64 & \textless{}0.001 & 1,905.30 & 97.5\% \\
 &  & \textsc{TER} & string & -9.91 & \textless{}0.001 & 1,911.39 & 93.2\% \\
 &  & \textsc{CER} & string & -7.11 & \textless{}0.001 & 1,925.33 & 92.3\% \\
 &  & \textsc{Rouge2} & string & 8.22 & \textless{}0.001 & 1,892.51 & 85.3\% \\
 &  & \textsc{BERTScore} & learned & 21.07 & \textless{}0.001 & 1,913.23 & 99.8\% \\
 &  & \textsc{Comet-QE} & learned & 3.58 & \textless{}0.001 & 1,924.70 & 64.5\% \\
 &  & \textsc{Comet} & learned & 11.96 & \textless{}0.001 & 1,918.07 & 96.1\% \\
 &  & \textsc{Bleurt20} & learned & 22.11 & \textless{}0.001 & 1,919.63 & 99.6\% \\
 &  & \textsc{Prism-QE} & learned & 13.43 & \textless{}0.001 & 1,916.54 & 99.9\% \\
 &  & \textsc{Prism} & learned & 25.55 & \textless{}0.001 & 1,924.74 & 100.0\% \\
\multirow{-14}{*}{\textbf{19}} &  \multirow{-14}{*}{grammar - interrogative} & BARTScore & learned & 6.37 & \textless{}0.001 & 1,925.93 & 96.0\% \\
\midrule
 &  & \textsc{Bleu} & string & 4.60 & \textless{}0.001 & 1,833.15 & 70.1\% \\
 &  & \textsc{Meteor} & string & 5.52 & \textless{}0.001 & 1,842.82 & 76.1\% \\
 &  & \textsc{ChrF} & string & 7.55 & \textless{}0.001 & 1,837.42 & 84.2\% \\
 &  & \textsc{ChrF2} & string & 6.93 & \textless{}0.001 & 1,837.52 & 82.6\% \\
 &  & \textsc{TER} & string & -2.26 & 0.024 & 1,835.00 & 53.7\% \\
 &  & \textsc{CER} & string & -3.00 & 0.003 & 1,827.77 & 68.7\% \\
 &  & \textsc{Rouge2} & string & 4.48 & \textless{}0.001 & 1,838.99 & 65.4\% \\
 &  & \textsc{BERTScore} & learned & 7.73 & \textless{}0.001 & 1,843.71 & 91.2\% \\
 &  & \textsc{Comet-QE} & learned & 4.09 & \textless{}0.001 & 1,842.82 & 80.7\% \\
 &  & \textsc{Comet} & learned & 6.04 & \textless{}0.001 & 1,843.56 & 91.5\% \\
 &  & \textsc{Bleurt20} & learned & 13.07 & \textless{}0.001 & 1,836.98 & 95.2\% \\
 &  & \textsc{Prism-QE} & learned & 3.86 & \textless{}0.001 & 1,841.41 & 90.4\% \\
 &  & PRISM & learned & 10.66 & \textless{}0.001 & 1,841.29 & 93.7\% \\
\multirow{-14}{*}{\textbf{20}} &  \multirow{-14}{*}{omission - adj/adv} & BARTScore & learned & 7.68 & \textless{}0.001 & 1,844.00 & 93.4\% \\
\midrule
 &  & \textsc{Bleu} & string & 4.13 & \textless{}0.001 & 1,917.46 & 61.7\% \\
 &  & \textsc{Meteor} & string & 4.70 & \textless{}0.001 & 1,927.87 & 63.8\% \\
 &  & \textsc{ChrF} & string & 6.85 & \textless{}0.001 & 1,923.46 & 80.7\% \\
 &  & \textsc{ChrF2} & string & 6.21 & \textless{}0.001 & 1,922.47 & 78.2\% \\
 &  & \textsc{TER} & string & -1.64 & 0.100 & 1,919.74 & 46.8\% \\
 &  & \textsc{CER} & string & -2.56 & 0.010 & 1,911.00 & 64.3\% \\
 &  & \textsc{Rouge2} & string & 3.62 & \textless{}0.001 & 1,921.73 & 55.3\% \\
 &  & \textsc{BERTScore} & learned & 16.81 & \textless{}0.001 & 1,926.62 & 96.5\% \\
 &  & \textsc{Comet-QE} & learned & 20.93 & \textless{}0.001 & 1,922.05 & 98.4\% \\
 &  & \textsc{Comet} & learned & 19.69 & \textless{}0.001 & 1,929.97 & 98.8\% \\
 &  & \textsc{Bleurt20} & learned & 30.30 & \textless{}0.001 & 1,830.05 & 98.8\% \\
 &  & \textsc{Prism-QE} & learned & 7.61 & \textless{}0.001 & 1,928.22 & 97.8\% \\
 &  & \textsc{Prism} & learned & 13.16 & \textless{}0.001 & 1,929.99 & 96.5\% \\
\multirow{-14}{*}{\textbf{21}} &  \multirow{-14}{*}{omission - content verb} & BARTScore & learned & 5.36 & \textless{}0.001 & 1,928.31 & 85.9\% \\
\midrule
 &  & \textsc{Bleu} & string & 5.42 & \textless{}0.001 & 1,926.09 & 71.2\% \\
 &  & \textsc{Meteor} & string & 7.52 & \textless{}0.001 & 1,942.42 & 77.5\% \\
 &  & \textsc{ChrF} & string & 9.81 & \textless{}0.001 & 1,938.83 & 88.9\% \\
 &  & \textsc{ChrF2} & string & 8.89 & \textless{}0.001 & 1,938.37 & 87.1\% \\
 &  & \textsc{TER} & string & -3.66 & \textless{}0.001 & 1,929.91 & 63.9\% \\
 &  & \textsc{CER} & string & -3.88 & \textless{}0.001 & 1,911.86 & 68.7\% \\
 &  & \textsc{Rouge2} & string & 4.86 & \textless{}0.001 & 1,938.91 & 64.1\% \\
 &  & \textsc{BERTScore} & learned & 20.07 & \textless{}0.001 & 1,941.62 & 97.9\% \\
 &  & \textsc{Comet-QE} & learned & 20.86 & \textless{}0.001 & 1,943.77 & 97.3\% \\
 &  & \textsc{Comet} & learned & 21.39 & \textless{}0.001 & 1,940.48 & 99.2\% \\
 &  & \textsc{Bleurt20} & learned & 34.88 & \textless{}0.001 & 1,811.22 & 99.2\% \\
 &  & \textsc{Prism-QE} & learned & 8.02 & \textless{}0.001 & 1,941.14 & 98.2\% \\
 &  & \textsc{Prism} & learned & 16.77 & \textless{}0.001 & 1,943.42 & 97.8\% \\
\multirow{-14}{*}{\textbf{22}} &  \multirow{-14}{*}{omission - noun} & BARTScore & learned & 9.61 & \textless{}0.001 & 1,933.99 & 90.2\% \\
\midrule
 &  & \textsc{Bleu} & string & 5.49 & \textless{}0.001 & 1,932.96 & 74.1\% \\
 &  & \textsc{Meteor} & string & 7.47 & \textless{}0.001 & 1,954.60 & 80.1\% \\
 &  & \textsc{ChrF} & string & 10.01 & \textless{}0.001 & 1,951.54 & 91.3\% \\
 &  & \textsc{ChrF2} & string & 9.47 & \textless{}0.001 & 1,949.00 & 90.5\% \\
 &  & \textsc{TER} & string & -3.84 & \textless{}0.001 & 1,942.84 & 67.3\% \\
 &  & \textsc{CER} & string & -4.87 & \textless{}0.001 & 1,926.02 & 72.6\% \\
 &  & \textsc{Rouge2} & string & 5.39 & \textless{}0.001 & 1,948.97 & 70.3\% \\
 &  & \textsc{BERTScore} & learned & 19.94 & \textless{}0.001 & 1,955.90 & 98.0\% \\
 &  & \textsc{Comet-QE} & learned & 16.41 & \textless{}0.001 & 1,955.97 & 94.7\% \\
 &  & \textsc{Comet} & learned & 18.65 & \textless{}0.001 & 1,951.01 & 98.5\% \\
 &  & \textsc{Bleurt20} & learned & 32.39 & \textless{}0.001 & 1,795.97 & 99.1\% \\
 &  & \textsc{Prism-QE} & learned & 8.17 & \textless{}0.001 & 1,945.91 & 96.0\% \\
 &  & \textsc{Prism} & learned & 18.64 & \textless{}0.001 & 1,949.93 & 98.3\% \\
\multirow{-14}{*}{\textbf{23}} &  \multirow{-14}{*}{omission - subject} & BARTScore & learned & 13.39 & \textless{}0.001 & 1,901.80 & 91.8\% \\
\midrule
 &  & \textsc{Bleu} & string & 6.06 & \textless{}0.001 & 1,336.10 & 80.4\% \\
 &  & \textsc{Meteor} & string & 9.32 & \textless{}0.001 & 1,351.99 & 94.4\% \\
 &  & \textsc{ChrF} & string & 11.63 & \textless{}0.001 & 1,351.88 & 97.5\% \\
 &  & \textsc{ChrF2} & string & 10.83 & \textless{}0.001 & 1,351.34 & 97.0\% \\
 &  & \textsc{TER} & string & -4.68 & \textless{}0.001 & 1,340.39 & 72.7\% \\
 &  & \textsc{CER} & string & -5.05 & \textless{}0.001 & 1,327.37 & 74.2\% \\
 &  & \textsc{Rouge2} & string & 6.20 & \textless{}0.001 & 1,348.43 & 79.8\% \\
 &  & \textsc{BERTScore} & learned & 21.78 & \textless{}0.001 & 1,351.75 & 98.5\% \\
 &  & \textsc{Comet-QE} & learned & 12.22 & \textless{}0.001 & 1,351.92 & 91.3\% \\
 &  & \textsc{Comet} & learned & 16.39 & \textless{}0.001 & 1,349.64 & 98.5\% \\
 &  & \textsc{Bleurt20} & learned & 32.33 & \textless{}0.001 & 1,288.20 & 99.4\% \\
 &  & \textsc{Prism-QE} & learned & 6.69 & \textless{}0.001 & 1,337.24 & 93.9\% \\
 &  & \textsc{Prism} & learned & 21.29 & \textless{}0.001 & 1,345.33 & 99.0\% \\
\multirow{-14}{*}{\textbf{24}} &  \multirow{-14}{*}{omission - named entry} & BARTScore & learned & 20.08 & \textless{}0.001 & 1,321.04 & 98.8\% \\

\bottomrule
\end{tabular}}
\end{center}
\label{tab:significance4}
\end{table*}

\begin{table*}
\small
\begin{center}
\resizebox{\columnwidth}{!}{\begin{tabular}{cllcrrrr} 
 \toprule
\multicolumn{1}{l}{} & & & \multicolumn{1}{l}{} & \multicolumn{3}{l}{Welsch \textit{t}-test} \\
\multicolumn{1}{l}{\multirow{-2}{*}{ID}}  & \multirow{-2}{*}{perturbation} & \multirow{-2}{*}{metric} & \multicolumn{1}{l}{\multirow{-2}{*}{type}} & \multicolumn{1}{l}{\textit{t}} & \multicolumn{1}{l}{\textit{p}-val} & \multicolumn{1}{l}{df}  & \multirow{-2}{*}{accuracy} \\

\midrule
 &  & \textsc{Bleu} & string & 5.04 & \textless{}0.001 & 1,744.49 & 67.7\% \\
 &  & \textsc{Meteor} & string & 5.27 & \textless{}0.001 & 1,754.73 & 69.6\% \\
 &  & \textsc{ChrF} & string & 3.77 & \textless{}0.001 & 1,753.94 & 84.1\% \\
 &  & \textsc{ChrF2} & string & 4.27 & \textless{}0.001 & 1,752.53 & 84.2\% \\
 &  & \textsc{TER} & string & -3.26 & 0.001 & 1,754.63 & 64.6\% \\
 &  & \textsc{CER} & string & -0.93 & 0.353 & 1,755.79 & 70.6\% \\
 &  & \textsc{Rouge2} & string & 5.78 & \textless{}0.001 & 1,744.41 & 64.1\% \\
 &  & \textsc{BERTScore} & learned & 21.29 & \textless{}0.001 & 1,748.10 & 99.8\% \\
 &  & \textsc{Comet-QE} & learned & 14.06 & \textless{}0.001 & 1,747.45 & 92.6\% \\
 &  & \textsc{Comet} & learned & 14.60 & \textless{}0.001 & 1,755.96 & 97.4\% \\
 &  & \textsc{Bleurt20} & learned & 14.96 & \textless{}0.001 & 1,750.64 & 97.6\% \\
 &  & \textsc{Prism-QE} & learned & 14.35 & \textless{}0.001 & 1,750.99 & 99.8\% \\
 &  & \textsc{Prism} & learned & 19.00 & \textless{}0.001 & 1,755.60 & 100.0\% \\
\multirow{-14}{*}{\textbf{25}} &  \multirow{-14}{*}{spelling - misspelled} & BARTScore & learned & 2.86 & 0.004 & 1,755.76 & 79.7\% \\
\midrule
 &  & \textsc{Bleu} & string & 5.69 & \textless{}0.001 & 1,976.50 & 68.1\% \\
 &  & \textsc{Meteor} & string & 5.52 & \textless{}0.001 & 1,996.82 & 66.4\% \\
 &  & \textsc{ChrF} & string & 3.48 & \textless{}0.001 & 1,995.51 & 85.8\% \\
 &  & \textsc{ChrF2} & string & 4.20 & \textless{}0.001 & 1,993.62 & 86.0\% \\
 &  & \textsc{TER} & string & -3.45 & \textless{}0.001 & 1,995.20 & 61.0\% \\
 &  & \textsc{CER} & string & -0.50 & 0.620 & 1,997.61 & 65.4\% \\
 &  & \textsc{Rouge2} & string & 6.55 & \textless{}0.001 & 1,980.84 & 61.8\% \\
 &  & \textsc{BERTScore} & learned & 19.73 & \textless{}0.001 & 1,993.08 & 99.5\% \\
 &  & \textsc{Comet-QE} & learned & 14.38 & \textless{}0.001 & 1,994.47 & 95.4\% \\
 &  & \textsc{Comet} & learned & 15.28 & \textless{}0.001 & 1,997.31 & 98.2\% \\
 &  & \textsc{Bleurt20} & learned & 16.73 & \textless{}0.001 & 1,987.94 & 98.7\% \\
 &  & \textsc{Prism-QE} & learned & 12.91 & \textless{}0.001 & 1,994.01 & 99.7\% \\
 &  & \textsc{Prism} & learned & 17.66 & \textless{}0.001 & 1,997.73 & 99.7\% \\
\multirow{-14}{*}{\textbf{26}} &  \multirow{-14}{*}{spelling - char removed} & BARTScore & learned & 3.38 & \textless{}0.001 & 1,997.57 & 80.1\% \\
\midrule
 &  & \textsc{Bleu} & string & 2.07 & 0.038 & 1,996.48 & 76.2\% \\
 &  & \textsc{Meteor} & string & 3.30 & \textless{}0.001 & 1,989.43 & 57.3\% \\
 &  & \textsc{ChrF} & string & 0.95 & 0.343 & 1,997.89 & 96.4\% \\
 &  & \textsc{ChrF2} & string & 1.96 & 0.050 & 1,997.73 & 98.3\% \\
 &  & \textsc{TER} & string & -2.64 & 0.008 & 1,993.32 & 55.8\% \\
 &  & \textsc{CER} & string & -0.81 & 0.420 & 1,997.92 & 80.3\% \\
 &  & \textsc{Rouge2} & string & 0.00 & 1.000 & 1,998.00 & 0.0\% \\
 &  & \textsc{BERTScore} & learned & 6.83 & \textless{}0.001 & 1,997.83 & 98.5\% \\
 &  & \textsc{Comet-QE} & learned & -6.33 & \textless{}0.001 & 1,987.30 & 12.0\% \\
 &  & \textsc{Comet} & learned & -1.10 & 0.273 & 1,997.74 & 39.8\% \\
 &  & \textsc{Bleurt20} & learned & 8.32 & \textless{}0.001 & 1,993.90 & 96.8\% \\
 &  & \textsc{Prism-QE} & learned & 6.29 & \textless{}0.001 & 1,995.83 & 99.9\% \\
 &  & \textsc{Prism} & learned & 9.01 & \textless{}0.001 & 1,997.79 & 100.0\% \\
\multirow{-14}{*}{\textbf{27}} &  \multirow{-14}{*}{punctuation - removed} & BARTScore & learned & 0.61 & 0.544 & 1,997.99 & 63.5\% \\
\midrule
 &  & \textsc{Bleu} & string & 5.01 & \textless{}0.001 & 1,977.93 & 90.3\% \\
 &  & \textsc{Meteor} & string & 5.67 & \textless{}0.001 & 1,996.17 & 69.4\% \\
 &  & \textsc{ChrF} & string & 2.61 & 0.009 & 1,995.62 & 97.9\% \\
 &  & \textsc{ChrF2} & string & 2.78 & 0.006 & 1,995.28 & 99.0\% \\
 &  & \textsc{TER} & string & -3.68 & \textless{}0.001 & 1,995.37 & 64.7\% \\
 &  & \textsc{CER} & string & -0.88 & 0.380 & 1,997.99 & 85.3\% \\
 &  & \textsc{Rouge2} & string & 0.00 & 1.000 & 1,998.00 & 0.0\% \\
 &  & \textsc{BERTScore} & learned & 15.85 & \textless{}0.001 & 1,983.20 & 99.3\% \\
 &  & \textsc{Comet-QE} & learned & 14.16 & \textless{}0.001 & 1,994.92 & 99.5\% \\
 &  & COMET & learned & 15.63 & \textless{}0.001 & 1,979.45 & 99.9\% \\
 &  & \textsc{Bleurt20} & learned & 16.35 & \textless{}0.001 & 1,997.90 & 99.3\% \\
 &  & \textsc{Prism-QE} & learned & 9.78 & \textless{}0.001 & 1,995.95 & 99.7\% \\
 &  & \textsc{Prism} & learned & 13.43 & \textless{}0.001 & 1,998.00 & 100.0\% \\
\multirow{-14}{*}{\textbf{28}} &  \multirow{-14}{*}{punctuation - added} & BARTScore & learned & 1.21 & 0.225 & 1,997.97 & 76.9\% \\
\midrule
 &  & \textsc{Bleu} & string & 1.44 & 0.149 & 1,997.07 & 18.6\% \\
 &  & \textsc{Meteor} & string & 9.42 & \textless{}0.001 & 1,978.96 & 84.0\% \\
 &  & \textsc{ChrF} & string & 0.00 & 1.000 & 1,998.00 & 0.0\% \\
 &  & \textsc{ChrF2} & string & 0.45 & 0.651 & 1,997.72 & 23.7\% \\
 &  & \textsc{TER} & string & -17.87 & \textless{}0.001 & 1,991.95 & 88.3\% \\
 &  & \textsc{CER} & string & -2.16 & 0.031 & 1,997.79 & 89.0\% \\
 &  & \textsc{Rouge2} & string & 0.14 & 0.888 & 1,998.00 & 1.2\% \\
 &  & \textsc{BERTScore} & learned & 19.16 & \textless{}0.001 & 1,995.29 & 99.8\% \\
 &  & \textsc{Comet-QE} & learned & 8.88 & \textless{}0.001 & 1,997.76 & 98.5\% \\
 &  & \textsc{Comet} & learned & 9.42 & \textless{}0.001 & 1,994.27 & 99.8\% \\
 &  & \textsc{Bleurt20} & learned & 14.75 & \textless{}0.001 & 1,985.71 & 99.5\% \\
 &  & \textsc{Prism-QE} & learned & 8.42 & \textless{}0.001 & 1,991.80 & 98.4\% \\
 &  & \textsc{Prism} & learned & 11.73 & \textless{}0.001 & 1,997.74 & 100.0\% \\
\multirow{-14}{*}{\textbf{29}} &  \multirow{-14}{*}{tokenized} & BARTScore & learned & 1.17 & 0.241 & 1,997.86 & 69.6\% \\
\midrule
 &  & \textsc{Bleu} & string & 14.04 & \textless{}0.001 & 1,955.69 & 87.8\% \\
 &  & \textsc{Meteor} & string & 0.00 & 1.000 & 1,998.00 & 0.0\% \\
 &  & \textsc{ChrF} & string & 11.23 & \textless{}0.001 & 1,990.65 & 90.6\% \\
 &  & \textsc{ChrF2} & string & 13.39 & \textless{}0.001 & 1,990.65 & 90.5\% \\
 &  & \textsc{TER} & string & 0.00 & 1.000 & 1,998.00 & 0.0\% \\
 &  & \textsc{CER} & string & -2.67 & 0.008 & 1,996.14 & 87.4\% \\
 &  & \textsc{Rouge2} & string & 0.00 & 1.000 & 1,998.00 & 0.0\% \\
 &  & \textsc{BERTScore} & learned & 25.36 & \textless{}0.001 & 1,957.60 & 99.3\% \\
 &  & \textsc{Comet-QE} & learned & 10.10 & \textless{}0.001 & 1,984.92 & 97.1\% \\
 &  & \textsc{Comet} & learned & 16.13 & \textless{}0.001 & 1,990.72 & 98.1\% \\
 &  & \textsc{Bleurt20} & learned & 22.51 & \textless{}0.001 & 1,997.97 & 99.3\% \\
 &  & \textsc{Prism-QE} & learned & 14.11 & \textless{}0.001 & 1,995.58 & 99.6\% \\
 &  & \textsc{Prism} & learned & 20.37 & \textless{}0.001 & 1,993.17 & 99.9\% \\
\multirow{-14}{*}{\textbf{30}} &  \multirow{-14}{*}{lowercase - whole} & BARTScore & learned & 7.04 & \textless{}0.001 & 1,997.80 & 93.6\% \\

\bottomrule
\end{tabular}}
\end{center}
\label{tab:significance5}
\end{table*}

\begin{table*}
\small
\begin{center}
\resizebox{\columnwidth}{!}{\begin{tabular}{cllcrrrr} 
 \toprule
\multicolumn{1}{l}{} & & & \multicolumn{1}{l}{} & \multicolumn{3}{l}{Welsch \textit{t}-test} \\
\multicolumn{1}{l}{\multirow{-2}{*}{ID}}  & \multirow{-2}{*}{perturbation} & \multirow{-2}{*}{metric} & \multicolumn{1}{l}{\multirow{-2}{*}{type}} & \multicolumn{1}{l}{\textit{t}} & \multicolumn{1}{l}{\textit{p}-val} & \multicolumn{1}{l}{df}  & \multirow{-2}{*}{accuracy} \\

\midrule
 &  & \textsc{Bleu} & string & 2.25 & 0.024 & 1,994.50 & 66.7\% \\
 &  & \textsc{Meteor} & string & 0.01 & 0.988 & 1,998.00 & 0.1\% \\
 &  & \textsc{ChrF} & string & 0.96 & 0.336 & 1,997.79 & 71.9\% \\
 &  & \textsc{ChrF2} & string & 1.74 & 0.082 & 1,997.24 & 72.0\% \\
 &  & \textsc{TER} & string & -0.01 & 0.992 & 1,998.00 & 0.0\% \\
 &  & \textsc{CER} & string & -0.70 & 0.482 & 1,997.86 & 71.4\% \\
 &  & \textsc{Rouge2} & string & 0.00 & 1.000 & 1,998.00 & 0.0\% \\
 &  & \textsc{BERTScore} & learned & 8.48 & \textless{}0.001 & 1,995.57 & 99.2\% \\
 &  & \textsc{Comet-QE} & learned & 4.82 & \textless{}0.001 & 1,998.00 & 95.0\% \\
 &  & \textsc{Comet} & learned & 3.96 & \textless{}0.001 & 1,997.61 & 96.3\% \\
 &  & \textsc{Bleurt20} & learned & 11.19 & \textless{}0.001 & 1,995.81 & 98.6\% \\
 &  & \textsc{Prism-QE} & learned & 5.05 & \textless{}0.001 & 1,997.98 & 98.9\% \\
 &  & \textsc{Prism} & learned & 7.03 & \textless{}0.001 & 1,997.96 & 99.0\% \\
\multirow{-14}{*}{\textbf{31}} &  \multirow{-14}{*}{lowercase - first word} & BARTScore & learned & 2.14 & 0.032 & 1,997.92 & 89.2\% \\
\midrule
 &  & \textsc{Bleu} & string & 66.14 & \textless{}0.001 & 999.00 & 100.0\% \\
 &  & \textsc{Meteor} & string & 135.10 & \textless{}0.001 & 999.08 & 100.0\% \\
 &  & \textsc{ChrF} & string & 166.71 & \textless{}0.001 & 1,000.01 & 100.0\% \\
 &  & \textsc{ChrF2} & string & 152.46 & \textless{}0.001 & 1,005.75 & 100.0\% \\
 &  & \textsc{TER} & string & -85.53 & \textless{}0.001 & 999.15 & 99.1\% \\
 &  & \textsc{CER} & string & -110.99 & \textless{}0.001 & 999.20 & 100.0\% \\
 &  & \textsc{Rouge2} & string & 89.55 & \textless{}0.001 & 999.00 & 99.9\% \\
 &  & \textsc{BERTScore} & learned & 103.38 & \textless{}0.001 & 1,643.56 & 100.0\% \\
 &  & \textsc{Comet-QE} & learned & 68.56 & \textless{}0.001 & 1,536.15 & 100.0\% \\
 &  & \textsc{Comet} & learned & 153.74 & \textless{}0.001 & 1,314.70 & 100.0\% \\
 &  & \textsc{Bleurt20} & learned & 386.70 & \textless{}0.001 & 1,296.35 & 100.0\% \\
 &  & \textsc{Prism-QE} & learned & 139.31 & \textless{}0.001 & 1,606.26 & 100.0\% \\
 &  & \textsc{Prism} & learned & 303.20 & \textless{}0.001 & 1,993.05 & 100.0\% \\
\multirow{-14}{*}{\textbf{32}} &  \multirow{-14}{*}{empty} & BARTScore & learned & 142.38 & \textless{}0.001 & 1,766.71 & 100.0\% \\
\midrule
 &  & \textsc{Bleu} & string & 62.67 & \textless{}0.001 & 1,001.22 & 100.0\% \\
 &  & \textsc{Meteor} & string & 119.40 & \textless{}0.001 & 1,113.17 & 100.0\% \\
 &  & \textsc{ChrF} & string & 122.77 & \textless{}0.001 & 1,087.45 & 100.0\% \\
 &  & \textsc{ChrF2} & string & 121.52 & \textless{}0.001 & 1,071.83 & 100.0\% \\
 &  & \textsc{TER} & string & -71.77 & \textless{}0.001 & 1,988.46 & 99.3\% \\
 &  & \textsc{CER} & string & -67.59 & \textless{}0.001 & 1,955.71 & 98.3\% \\
 &  & \textsc{Rouge2} & string & 88.50 & \textless{}0.001 & 1,011.66 & 99.9\% \\
 &  & \textsc{BERTScore} & learned & 184.64 & \textless{}0.001 & 1,947.30 & 100.0\% \\
 &  & \textsc{Comet-QE} & learned & 3.26 & 0.001 & 1,975.84 & 55.6\% \\
 &  & COMET & learned & 81.20 & \textless{}0.001 & 1,904.40 & 100.0\% \\
 &  & \textsc{Bleurt20} & learned & 251.23 & \textless{}0.001 & 1,978.21 & 100.0\% \\
 &  & \textsc{Prism-QE} & learned & 87.09 & \textless{}0.001 & 1,556.55 & 94.8\% \\
 &  & \textsc{Prism} & learned & 191.18 & \textless{}0.001 & 1,997.16 & 100.0\% \\
\multirow{-14}{*}{\textbf{33}} &  \multirow{-14}{*}{different} & BARTScore & learned & 147.07 & \textless{}0.001 & 1,732.24 & 100.0\% \\
\midrule
 &  & \textsc{Bleu} & string & 55.60 & \textless{}0.001 & 1,047.83 & 100.0\% \\
 &  & \textsc{Meteor} & string & 55.33 & \textless{}0.001 & 1,519.93 & 99.4\% \\
 &  & \textsc{ChrF} & string & 43.90 & \textless{}0.001 & 1,592.64 & 100.0\% \\
 &  & \textsc{ChrF2} & string & 45.40 & \textless{}0.001 & 1,515.48 & 100.0\% \\
 &  & \textsc{TER} & string & -57.36 & \textless{}0.001 & 1,595.93 & 99.1\% \\
 &  & \textsc{CER} & string & -63.66 & \textless{}0.001 & 1,288.57 & 96.9\% \\
 &  & \textsc{Rouge2} & string & 75.75 & \textless{}0.001 & 1,198.84 & 99.9\% \\
 &  & \textsc{BERTScore} & learned & 134.81 & \textless{}0.001 & 1,995.55 & 100.0\% \\
 &  & \textsc{Comet-QE} & learned & 98.03 & \textless{}0.001 & 1,962.18 & 100.0\% \\
 &  & \textsc{Comet} & learned & 111.85 & \textless{}0.001 & 1,910.48 & 100.0\% \\
 &  & \textsc{Bleurt20} & learned & 128.11 & \textless{}0.001 & 1,982.63 & 100.0\% \\
 &  & \textsc{Prism-QE} & learned & 106.85 & \textless{}0.001 & 1,838.50 & 100.0\% \\
 &  & \textsc{Prism} & learned & 140.83 & \textless{}0.001 & 1,847.24 & 100.0\% \\
\multirow{-14}{*}{\textbf{34}} &  \multirow{-14}{*}{unintelligible (shuffled)} & BARTScore & learned & 65.65 & \textless{}0.001 & 1,828.42 & 100.0\% \\
\midrule
 &  & \textsc{Bleu} & string & -90.34 & \textless{}0.001 & 999.00 & 100.0\% \\
 &  & \textsc{Meteor} & string & -60.37 & \textless{}0.001 & 999.00 & 100.0\% \\
 &  & \textsc{ChrF} & string & -76.50 & \textless{}0.001 & 999.00 & 100.0\% \\
 &  & \textsc{ChrF2} & string & -78.24 & \textless{}0.001 & 999.00 & 100.0\% \\
 &  & \textsc{TER} & string & 63.71 & \textless{}0.001 & 999.00 & 100.0\% \\
 &  & \textsc{CER} & string & 56.77 & \textless{}0.001 & 999.00 & 100.0\% \\
 &  & \textsc{Rouge2} & string & -77.52 & \textless{}0.001 & 999.00 & 100.0\% \\
 &  & \textsc{BERTScore} & learned & -61.70 & \textless{}0.001 & 999.00 & 100.0\% \\
 &  & \textsc{Comet-QE} & learned & 1.43 & 0.152 & 1,997.61 & 44.4\% \\
 &  & \textsc{Comet} & learned & -25.94 & \textless{}0.001 & 1,983.70 & 100.0\% \\
 &  & \textsc{Bleurt20} & learned & -94.47 & \textless{}0.001 & 1,160.41 & 100.0\% \\
 &  & \textsc{Prism-QE} & learned & 7.59 & \textless{}0.001 & 1,991.89 & 14.3\% \\
 &  & \textsc{Prism} & learned & -50.57 & \textless{}0.001 & 1,159.77 & 99.4\% \\
\multirow{-14}{*}{\textbf{35}} &  \multirow{-14}{*}{reference} & BARTScore & learned & -38.54 & \textless{}0.001 & 1,248.78 & 99.8\% \\

\bottomrule

\end{tabular}}
\end{center}
\caption{A two-samples Welsch \textit{t}-test is conducted on each metric to compare \texttt{SCORE}$(r, t)$ and \texttt{SCORE}$(r, t')$ (see Section~\ref{sec.using.translation.perturbations.to.diagnose.MT}) of each perturbation type. The tests are implemented in Python using the package \texttt{scipy~\citep{2020SciPy-NMeth}}. Degrees of Freedom (DF) are estimated using the Welch-Satterthwaite equasion for Degrees of Freedom. The accuracy on the baseline perturbation \emph{35} (reference as translation) was reversed, as one can expect the metric to prefer translation identical with the reference.}
\label{tab:significance5}
\end{table*}

\end{document}